\documentclass{article}

\usepackage[preprint]{neurips_2022}

\usepackage{graphicx}

\usepackage{float}
\usepackage{tablists}
\usepackage{xcolor}
\usepackage{tikz}

\usepackage{pdfpages}

\usepackage{caption}
\usepackage[explicit]{titlesec}

\usepackage{subfig}

\usepackage{amsmath}
\usepackage{amssymb}

\usepackage{times}  

\title{Why not both? Complementing explanations with uncertainty, and the role of self-confidence in Human-AI collaboration}

\author{%
  Ioannis Papantonis\\
  School of Informatics\\
  The University of Edinburgh\\
  Edinburgh, UK \\
  \texttt{i.papantonis@sms.ed.ac.uk} \\
  \And
  Vaishak Belle \\
  School of Informatics\\
  The University of Edinburgh \\
  Edinburgh, UK \\
  \texttt{vbelle@ed.ac.uk} \\
}

\begin{document}

\maketitle

\begin{abstract}
AI and ML models have already found many applications in critical domains, such as healthcare and criminal justice. However, fully automating such high-stakes applications can raise ethical or fairness concerns.  Instead, in such cases, humans should be assisted by automated systems so that the two parties reach a joint decision, stemming out of their interaction. In this work we conduct an empirical study to identify how uncertainty estimates and model explanations affect users' reliance, understanding, and trust towards a model, looking for potential benefits of bringing the two together.
Moreover, we seek to assess how users' behaviour is affected by their own self-confidence in their abilities to perform a certain task, while we also discuss how the latter may distort the outcome of an analysis based on agreement and switching percentages.
\end{abstract}

\section{Introduction}

AI and ML models have already become an indispensable component in many applications, ranging from medical diagnosis to criminal justice. However, full automation is not always desirable, especially in high-stakes applications, for example due to ethical \citep{naik2022legal} or fairness \citep{10.1145/3457607} concerns. Instead, in such cases, humans should be assisted by  automated systems so that the two parties reach a joint decision, stemming out of their interaction. The advantage of this approach is that while it makes use of sophisticated AI systems, humans retain full agency over the final decision, limiting the adverse effect of potential poor model predictions. One of the primary objectives of this human-AI collaboration is to achieve high performance, a goal that requires human users to  be able to decide when to follow the model’s predictions, which is a multi-faceted objective, influenced by complex interactions between multiple factors \citep{lee2004trust,hoff2015trust,adams2003trust}. 

Identifying such factors as well as the way they influence user behaviour and attitude towards a model has been an active research area for decades within the human factors and the AI communities, resulting in several behavioural theories describing the dynamics of the human-AI interaction \citep{lee1992trust,linegang2006human,madsen2000measuring}. A consistent point of convergence among these theories is that both model-related factors, such as the extent to which a model is perceived to be \textit{reliable} and \textit{understandable}, and user-related factors, such as their \textit{self-confidence} in their abilities to carry out a task, play a crucial role in the formation of the human-AI relationship. 


As far as model-related factors are concerned, the emergence of explainable AI has sparked a surge of empirical studies that explore the effect of different explanation styles on model understanding, or the capacity of explanations to allow users detect unfair model behaviour \citep{lai2019human,wang2021explanations,dodge2019explaining,lai2020chicago}. Moreover, with respect to reliability, recent studies have contrasted the influence of model predictions, uncertainty estimates, and explanations on users' perceived model reliability, comparing their relative effectiveness  on  instilling trust and/or inducing a complementary performance benefit, where the joint human-AI accuracy is superior to the individual accuracy of either party \citep{zhang2020effect,bansal2021does,green2019principles,lundberg2018explainable}. While this is an ongoing endeavour, there has been substantial evidence suggesting that uncertainty estimates are at least as effective as explanations in achieving these goals. Moreover, uncertainty estimates are arguably simpler to implement and communicate to diverse audiences, raising questions about the overall utility of explanations.



Having said that, surveys that consider both uncertainty estimates and explanations, usually view them as competing sources of reliability-related information. While this approach has the merit of providing a common ground upon which it is possible to compare the two, it reduces explanations to reliability indicators, even though their primarily function is to enhance understanding \citep{hoffman2018metrics}. In addition, while prior research suggests that information regarding reliability and understanding have complementary functions \citep{zuboff1988age,sheridan1989trustworthiness,lee1992trust,madsen2000measuring,kelly2003guidelines}, the aforementioned approach fails to capture this aspect and provide relevant insights. For example, uncertainty estimates may help users decide the extent to which to rely on a model, but they provide no justifications in cases where a model makes incorrect predictions, hindering model acceptance \citep{ashoori2019ai}. On the other hand, while explanations mitigate this issue, inferring a model’s prediction and uncertainty based on explanations alone, requires substantial technical expertise, while also inducing a very high  cognitive load, making it an inefficient strategy for practical applications \citep{kaur2020interpreting}.


In addition to the above, users' self-confidence in their abilities to complete a certain task is another factor that influences multiple aspects of the human-AI relationship \citep{lee1992trust,lewandowsky2000dynamics,de2003effects,lee2004trust}. A number of empirical surveys have studied this effect in tasks where humans function as operators, deciding whether to perform a task manually or allocate it to a model, providing evidence that humans' self-confidence has a significant influence on trust and reliance. Despite such findings, in the context of joint decision-making, where humans are always in charge of taking a decision, and the model takes on an advisory role, self-confidence has received very little attention. This leaves a significant gap in empirical investigations, especially considering that many surveys in the domain explore questions concerning trust in automation.

Moreover, another point that warrants further consideration is the way trust is operationalized in recent surveys. In particular, trust is almost exclusively assessed through the lens of \textit{agreement} and \textit{switching percentages} \citep{zhang2020effect}, as opposed to using specialized trust measuring scales, such as those developed in \citep{madsen2000measuring,jian2000foundations,adams2003trust,cahour2009does}. Nevertheless, it is well established that both of these percentages measure reliance, not trust, and that they may fail to account for confounding variables, such as time constraints, inherent application risks, or users' own self-confidence \citep{miller2016behavioral,chancey2013role}. This is because although trust has been identified to mediate reliance on automation, trust is a broader attitude towards automation, while reliance is a behaviour that may potentially constitute a manifestation of trust \citep{ajzen1980understanding,lee2004trust}. For example, it is possible for one to rely on a model without really trusting it, simply because one lacks the background to take an informed decision. On the other hand, it is also possible for users  to base their decisions solely on their own knowledge, so any agreement with the model is only coincidental, not a manifestation of reliance or trust.

In this work we attempt to address these issues by conducting an empirical study to identify the effect of self-confidence and various types of model assistance on human-AI collaboration. In particular, we seek to identify how the joint accuracy of the ensemble is affected by users' confidence, as well as whether there are differences in user behaviour depending on the provided level of model assistance. Moreover, we seek to uncover potential non accuracy-related benefits of bringing together uncertainty estimates and explanations, looking for differences in terms of reliance, understanding, and trust towards the model. With this we aim to provide evidence that although uncertainty estimates may be as effective as explanations with respect to performance, the latter influence other key aspects, so pairing the two together induces a complementary effect by leveraging the simplicity of uncertainty estimates and the unique insights offered by explanations. More specifically we present the following contributions:

\begin{itemize}
    \item We design and implement an online empirical study with human participants.
    \item We identify a complementary effect between uncertainty estimates and explanations, with the former being sufficient for improving performance, and the latter leading to significant improvements in both subjective and objective model understanding.
    \item We provide evidence that human self-confidence significantly influences the joint human-AI accuracy, while we also illustrate the pitfalls of not properly adjusting for this effect.
    \item We showcase how different uncertainty measures influence user behaviour.
    \item We show that both human and model confidence affect reliance, understanding, and trust.
    \item We demonstrate the limitations of using switching and agreement percentages as a proxy for trust.
    
\end{itemize}



\section{Related work}

The importance of establishing a transparent relationship between human users and automation on fostering  an effective collaboration between the two parties  has been consistently identified in prior literature. In \citep{bhatt2021uncertainty} the authors called for utilizing diverse estimates that convey multiple aspects of the underlying model uncertainty  to promote transparency and help users comprehend the degree to which a model’s predictions should be followed. Moreover, the findings in \citep{ashoori2019ai} suggested that in high-stakes applications uncertainty estimates might not be enough, since the absence of explanations may lead to users entirely dismissing a model, regardless of its accuracy.

Motivated by such discussions, a growing body of recent empirical investigations focus on the relative effect of uncertainty and explanations on joint  accuracy and trust. For example, the findings in \citep{zhang2020effect}, suggested that simply providing participants with information about model confidence, i.e. the probability a model assigns to its predictions, is more effective than explanations in improving trust and joint accuracy, as well as that explanations were not successful in allowing  participants disentangle between high and low confidence predictions. Moreover, the results in \citep{lai2019human} demonstrated that the best joint accuracy was achieved when presenting information containing the model’s prediction paired with the corresponding model confidence, in line with \citep{zhang2020effect}. Pairing local feature importance explanations and model predictions was slightly less effective, while presenting explanations alone, led only to a minor improvement compared to the baseline.

Another related study is presented in \citep{bansal2021does}, which explores whether combining model confidence and explanations can further improve the accuracy of the human-AI team. The resulting analysis showed that when both parties had comparable individual accuracy, then presenting participants with the model’s prediction and confidence led to the ensemble achieving superior joint accuracy. The authors found no further improvement when pairing this information with explanations, concluding that the former strategy is as effective as the latter, while also being substantially simpler.

Moreover, in both \citep{bansal2021does,zhang2020effect} it has been acknowledged  that a user’s self-confidence should have an effect on the joint human-AI accuracy in the context of decision-making tasks. Despite this being an intuitive remark, to the best of our knowledge, this idea has not been empirically verified. This is in contrast to alternative settings, such as when humans function as operators being in charge of deciding whether to perform a task manually or delegate it to a model, where the role of self-confidence has received considerable attention. In \citep{lee1994trust}, the authors provided evidence that participants turned into automation only when their trust in its capabilities exceeded their own, otherwise they tended to performed a task manually. These findings are extended in \citep{de2003effects}, where the results indicated that there exists a fundamental bias towards people trusting their own abilities, instead of the model. Moreover, in \citep{lewandowsky2000dynamics}, it is shown that participants’ self-confidence determined whether they retained control  or not, strengthening the previous findings.

Apart from exploring the effect of explanations on accuracy in decision-making tasks, other surveys focus on alternative questions, such as the one in \citep{dodge2019explaining}, which explored the efficacy of explanations in helping human users detect unfair model behaviour. Interestingly, the results revealed that local explanations were the most effective in exposing  fairness discrepancies among individuals, while global ones instilled more confidence in the users that  their understanding was correct. In addition, the study in \citep{wang2021explanations}, brought  a new perspective by exploring the comparative effect of explanation styles on model understanding, across datasets of varying difficulty. The final results uncovered that the difficulty of the application significantly influenced the effect of explanations on model understanding, while also indicating that local explanations improved participants’ objective understanding, and that global explanations improved their self-reported subjective understanding.

Finally, a methodological approach shared by many recent studies on trust in the human-AI collaboration, is that trust is measured using agreement and switching percentages. The former is the fraction of times that the user and the model agreed on their final predictions, while the latter is the percentage of times users switched their predictions to follow the model, assuming the two parties initially disagreed. This approach is in contrast with the predominant practice in the human factors and human-computer interaction communities, where trust is assessed based on either specialized trust measuring scales such as \citep{madsen2000measuring,jian2000foundations,adams2003trust,cahour2009does}, sophisticated implicit behavioural measures \citep{de2003effects,miller2016behavioral}, or combinations thereof. Furthermore, focusing exclusively on the aforementioned percentages to measure trust poses a major methodological shift, since both of them are indicators of reliance \citep{miller2016behavioral,lee2004trust}, so this paradigm presupposes that trust can be indirectly inferred through reliance. 
 

\section{Study Overview}


In this study we design a salary prediction task, and we seek to answer questions along two principal axes. On the one hand, we follow the discussions in \citep{bansal2021does} and we seek to obtain deeper insights regarding the role of the interaction of human and model confidence in influencing joint accuracy. As the authors note, this interaction should play an important part in regulating  joint accuracy, however, there is no concrete evidence supporting this view. In this work we fill this gap, while further expanding on this idea, exploring not only how the joint accuracy is affected, but also how reliance, understanding, and trust are shaped as a result of this interaction. On the other hand, we seek to find evidence of added benefits of pairing uncertainty estimates with explanations. Recent surveys have consistently demonstrated that in terms of accuracy the former is at least as effective as the latter, suggesting that uncertainty alone is enough to promote an effective human-AI collaboration. However, it is still unclear whether combining uncertainty and explanations can yield alternative, non accuracy related benefits. In this work we look for differences with respect to model understanding, which is an important factor, linked to aspects such as model acceptance and long-term adoption \citep{adams2003trust}. Similarly, motivated by the discussions in \citep{bhatt2021uncertainty}, we explore the effect of combining uncertainty measures of different scope on users’ behaviour. In particular, we ask the following research questions:

   
\begin{itemize}
    \item[\textbf{RQ1}] How is joint predictive performance influenced by the interaction of human confidence, model confidence, and the degree of model assistance?
    
    \item[\textbf{RQ2}] How are reliance, understanding, and trust towards the model affected by the same factors?
    
    \item[\textbf{RQ3}] Does the combination of explanations and uncertainty measures offer non accuracy-related, complementary benefits?

    \item[\textbf{RQ4}] How uncertainty estimates of varying scope influence user behaviour? 
\end{itemize}


Studying these questions, we aim to assess the role of self-confidence in decision-making tasks, as well as how different combinations of information elicit differences in user behaviour. Moreover, we  demonstrate the pitfalls of using switching and agreement percentages as a proxy for studying trust. In sum, our goal is to uncover concrete advantages  of employing combinations of diverse information sources, promoting research that further expands on this topic. This is especially important considering that in naturalistic settings, stakeholders expect combinations of multiple sources of information. More specifically, we aim to test the following hypotheses:

\begin{itemize}

    \item[\textbf{H1}] Superior joint accuracy will be observed when humans have low self-confidence, and the model makes high confidence predictions. Moreover, pairing model prediction and confidence will be sufficient to induce this effect.
    \item[\textbf{H2}] Participants provided with explanations will have better model understanding.
    \item[\textbf{H3}] Reliance, understanding and trust towards the model will be affected by both human confidence and model confidence, as follows:
    \begin{itemize}
        \item[{\small \textbf{H3.1}}] Reliance will be affected primarily by human confidence, and to a lesser extent by model confidence. Furthermore,   we expect to find an increase in reliance when humans have low confidence and the model makes high confidence predictions .
        \item[{\small \textbf{H3.2}}] Understanding will be similarly affected by both human and model confidence . In addition,  we expect an increase in understanding when both parties have high confidence.
        \item[{\small \textbf{H3.3}}] Trust will be affected primarily by human confidence, and to a lesser extent by model confidence. We also expect an increase in trust when both parties have high confidence.
    \end{itemize}
    
    \item[\textbf{H4}] The difference between uncertainty measures of distinct scopes (global vs local) will induce differences in user behaviour.

\end{itemize}


    
    



\subsection{Experimental Design}

\subsubsection{Participants}

We recruited $112$ participants from Amazon Mechanical Turk for our experiment. $49$ participants were women, and $63$ were men. $18$ participants were between age $18$ and $29$, $45$ between age $30$ and $39$, $23$ between $40$ and $49$, and $26$ were over $50$ years old. Furthermore, our task was available only to USA residents, due to the fact that the selected dataset contained information that was relevant to the USA social context.

\subsubsection{Dataset}
We designed a modified version of the task presented in \citep{zhang2020effect}, where participants had to predict whether a person's annual salary was greater than $50000$ dollars. However, since this task was based on the Adult dataset, which contains data from the $1994$ Census\footnote{Link: https://archive.ics.uci.edu/ml/datasets/adult}, we needed to adjust the salary threshold to account for inflation. Considering that in this time span the US dollar has seen a cumulative price increase of $101.09\%$, the adjusted value became $100500$, which was rounded to $100000$ dollars. The dataset contains $48842$ instances, and each one is comprised of $14$ features. Following the authors in \citep{zhang2020effect}, we opted for using only the $8$ most relevant ones, so participants were not overloaded with information. These features corresponded to a person's: age, employer, education, marital status, occupation, ethnic background, gender, as well as the hours-per-week spent working. We trained a gradient boosting decision tree model on $80\%$ of this dataset, leaving the remaining $20\%$ to test its final performance, which turnout out to be $82\%$.

\subsubsection{Task Instances}

The reason we selected the Adult dataset, was that it contains instances of varying difficulty, where some of them are relatively easy to predict for lay users with no prior related experience, while others can be significantly harder. This allows for actively manipulating participants’
self-confidence to study its effect on various aspects of the human-AI collaboration. In contrast, the authors in \citep{wang2021explanations}, utilized datasets that were either relatively easy or very hard for lay users. A limitation of this approach is that when participants perform a task for which they have no related knowledge or intuition, they are in a state of absolute ignorance promoting a blind reliance on the model, which is first very different from being uncertain and second very far from real-life situations. Indeed, it is highly improbable that a model will be employed by stakeholders having no knowledge/intuition regarding the application at hand. Instead, our prediction task is quite intuitive and mostly requires common sense knowledge, while also allowing for different degrees of confidence in one's predictions.

In order to select the actual task instances, we first set the threshold for low confidence model predictions at $65\%$, meaning that any prediction with probability not exceeding that number, was considered to be a low confidence model prediction. The corresponding threshold for high confidence predictions was set at $80\%$. We intentionally opted for a relatively large gap between the two thresholds in order to avoid the interval in-between where it is ambiguous whether a prediction should be seen as having low or high confidence. We then went through the resulting filtered dataset looking for instances of varying complexity, from a human’s perspective. After completing this step, we needed to make sure that humans and model have comparable individual performances, to match the setting in \citep{bansal2021does}. Following the suggestion in \citep{zhang2020effect}, we used a stratified sampling approach, constraining the model accuracy to be $75\%$, since the unconstrained accuracy ($82\%$) was very high for lay people. By the end of this procedure, we identified $56$ instances, equally divided into the $4$ configurations of human/model confidence: (Human - High \& Model - High), (Human - High \& Model - Low), (Human - Low \& Model - High), (Human - Low \& Model - Low). In order to verify that these instances were indeed effective both in inducing different states of human confidence and in allowing for comparable human-model performance, we recruited $15$ participants from Amazon Mechanical Turk, asking them to provide  a confidence score and prediction  for each of  these datapoints. Finally, we confirmed that our categorization was effective at inducing a different level of self-confidence to lay users ($Z=8$, $p<0.001$), as well as that the selected instances allowed for a comparable accuracy between participants and model (Average human accuracy $=65.6\%$, $95\%$ confidence interval $(54.2, 76.4)$)\footnote{For the former we used Wilcoxon's signed-rank test, while the latter was estimated using the bootstrap method.}.

\subsubsection{Design}\label{design1}

\begin{figure}[htp]
\centering
\subfloat[Unassisted prediction]{\label{trial1}
\centering
  \includegraphics[width=0.6\columnwidth]{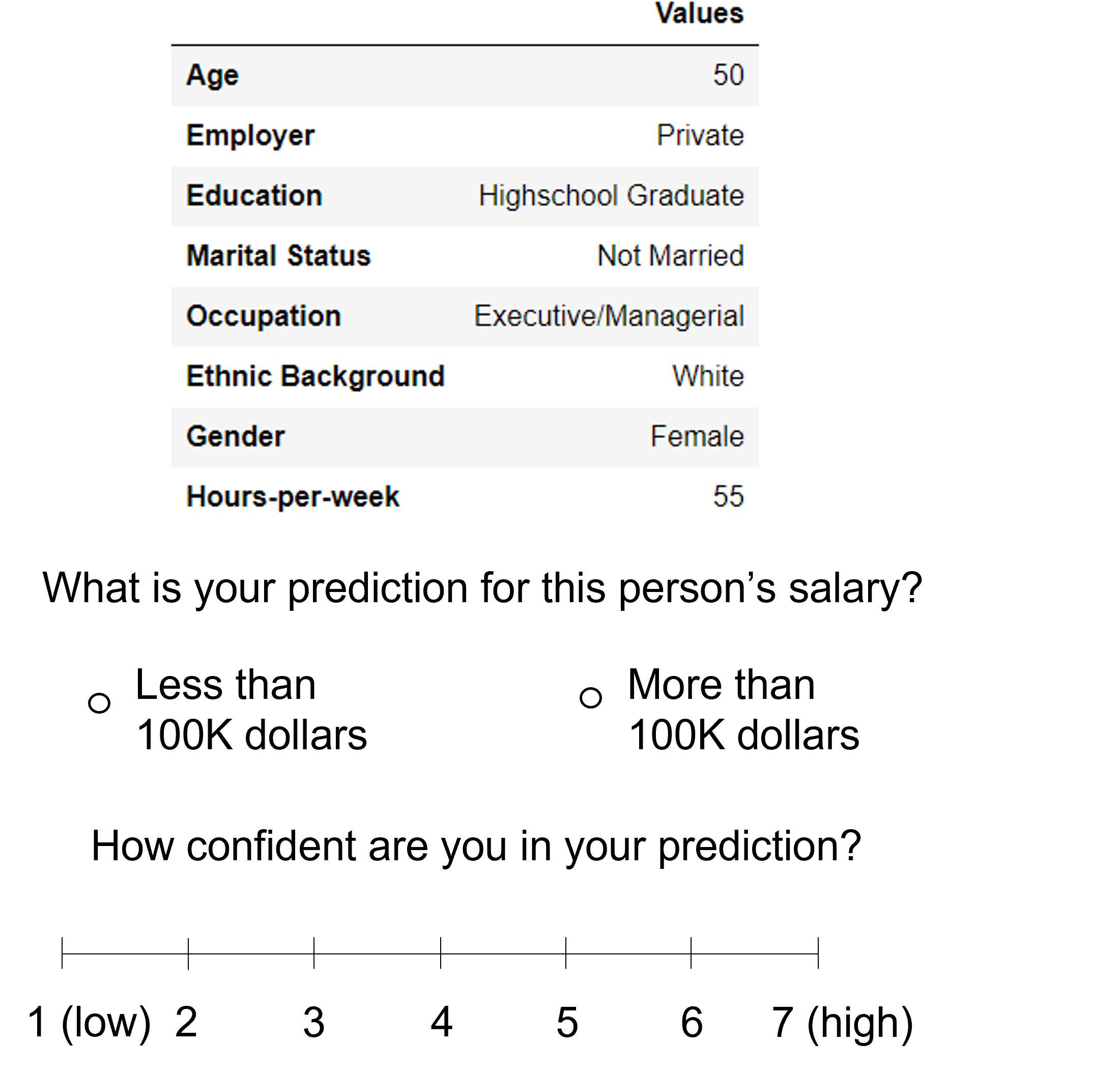}%
}
\hspace{60pt}
\centering
\subfloat[Assisted prediction]{\label{trial2}
\centering
  \includegraphics[width=\columnwidth]{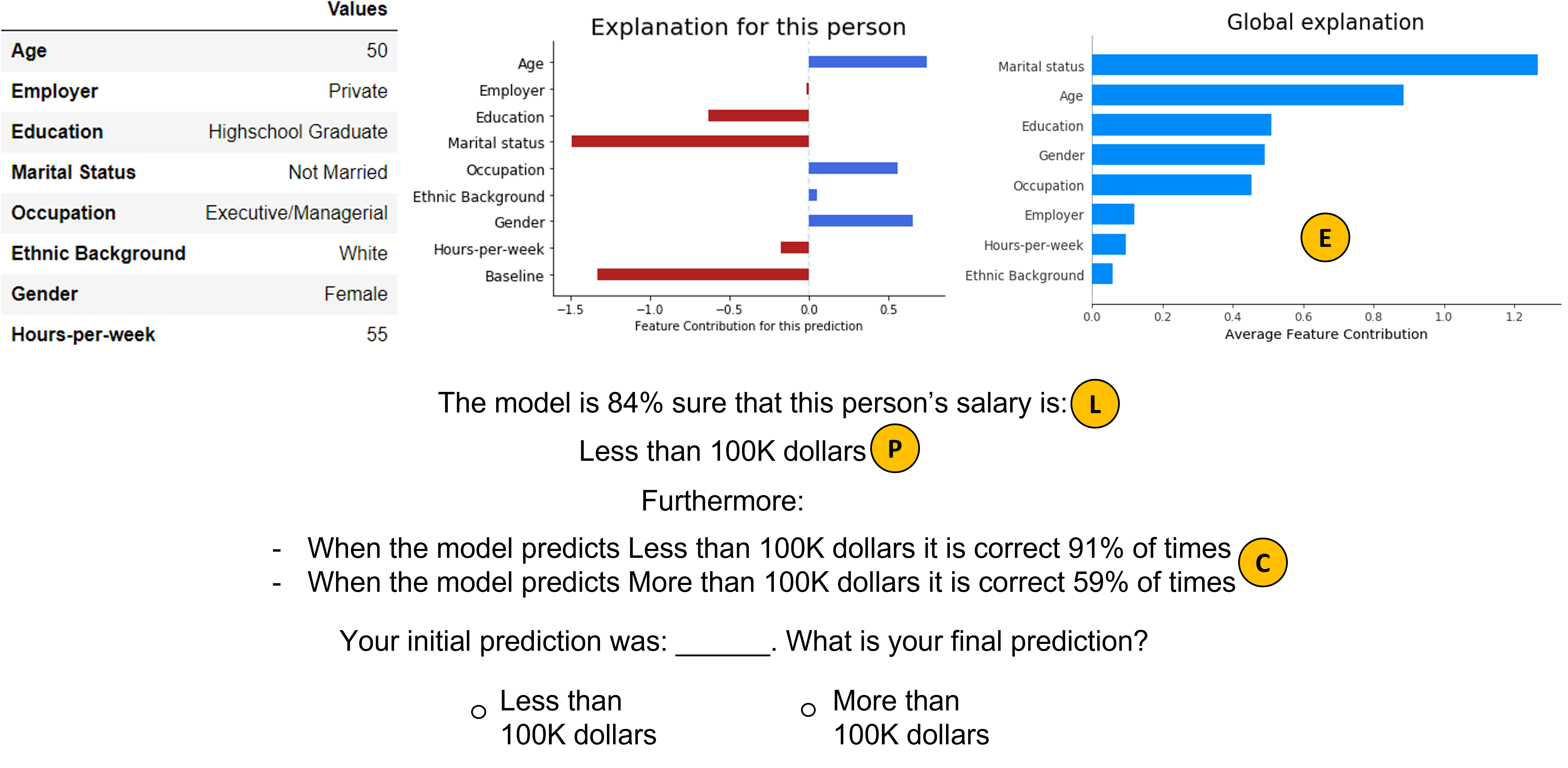}%
}

\caption{(a) Participants needed to inspect a datapoint and provide their unassisted prediction/confidence. (b) The model assistance presented to participants, depending on condition. Participants in the \textbf{Prediction} condition were shown the datapoint and the information next to \begin{tikzpicture}\protect \draw[fill=orange!40!yellow] (1ex,1ex) circle [radius=1.5ex] node {\textbf{P}};\end{tikzpicture}. In \textbf{Local}, they were shown \begin{tikzpicture}\protect \draw[fill=orange!40!yellow] (0.01ex,0.01ex) circle [radius=1.5ex] node {\textbf{P}}; \protect \draw[draw=white,fill=white]  (2.5ex,0.01ex) circle [radius=0.5ex] node {+}; \protect \draw[fill=orange!40!yellow] (5ex,0.01ex) circle [radius=1.5ex] node {\textbf{L}};\end{tikzpicture}. In \textbf{Combined}, they were shown \begin{tikzpicture}\protect \draw[fill=orange!40!yellow] (0.01ex,0.01ex) circle [radius=1.5ex] node {\textbf{P}}; \protect \draw[draw=white,fill=white]  (2.5ex,0.01ex) circle [radius=0.5ex] node {+}; \protect \draw[fill=orange!40!yellow] (5ex,0.01ex) circle [radius=1.5ex] node {\textbf{L}}; \protect \draw[draw=white,fill=white]  (7.5ex,0.01ex) circle [radius=0.5ex] node {+}; \protect \draw[fill=orange!40!yellow] (10ex,0.01ex) circle [radius=1.5ex] node {\textbf{C}};\end{tikzpicture}. Finally, in \textbf{Explanations}, they were shown all the information contained in this slide. }
\label{trials}

\end{figure}

In order to address our research questions, we designed a prediction task where in each trial participants needed to go through a three-step process. First, they had to inspect an instance, and provide an initial prediction about that person’s salary, as well as an estimate of their confidence. Following that they were provided with varying levels of model assistance, depending on the condition (see below), and then they were asked to give their final prediction, where they were free to either maintain or change their initial one. Figure \ref{trials} shows an example of this procedure. Finally, participants needed to provide an estimate of how much they relied on the model for that prediction, how much they felt they understood its decision-making process, as well as to which extent they trusted the model’s prediction. These three steps were repeated in each trial, and after completing the task participants were given a test comprised of $9$ multiple choice questions, adapted from \citep{wang2021explanations}, to assess their objective understanding of the model.

In more detail, there were $4$ experimental conditions, each one providing an increasing level of model support:

\begin{itemize}
    \item \textbf{Prediction:} In this condition, after participants submitted their initial prediction and confidence score, they are shown only the model's prediction for the same instance. After inspecting it, they are asked to submit their final answer. This serves as the baseline condition, providing only minimal model assistance. 
    \item \textbf{Local Confidence:} In this condition, participants were shown both the model's prediction and the corresponding model confidence, i.e the probability that the model assigned to each prediction. 

\item \textbf{Combined Confidence:} In this condition, participants  were shown all the information that was available in the previous one, plus the recall for each class, , i.e. the fraction of times an instance is correctly identified by the model as being a member of the class.  Here, recall acts as a global meta uncertainty estimate, providing information about the robustness of a model's own confidence. Combining these uncertainty measures should help participants gain a more refined picture of the model's performance, since knowing that a model is, say, $80\%$ confident in its prediction, but predictions for this class are correct only $50\%$ of the time, is more informative  than just knowing the model's confidence.
    
 
 \item \textbf{Explanations:} In this condition, participants were shown all of the previous information, as well as a local and a global explanation. Based on the findings in \citep{wang2021explanations}, we employed feature importance explanations for both, due to their effectiveness in promoting a better model understanding. 
 Local explanations showed how much each feature influenced the model to reach a particular prediction, while global ones displayed the average overall impact of each feature. All explanations were generated based on SHAP. 
 
\end{itemize}


Participants were randomly assigned to one of the four conditions. Within subjects we manipulated model confidence and human confidence, such that participants in each condition were presented with an equal number of trials with each confidence combination. More precisely, each participant was presented with $4$ instances of each of the following certainty combinations: (Human - High \& Model - High), (Human - High \& Model - Low), (Human - Low \& Model - High), (Human - Low \& Model - Low). Participants were also asked to provide their confidence in each of their predictions, which was used to confirm that our manipulation was successful in inducing varying degrees of confidence in this sample too ($Z=200$, $p<0.001$).

In addition, we matched the number of instances with people earning less/more than $100$K dollars within each certainty combination, such that two out of the four instances of each combination showed people gaining more than $100$K dollars. Order of presentation of instances was random. Our dependent variables are accuracy, reliance, subjective understanding of the model, trust and objective understanding of the model.

\subsubsection{Procedure}

Upon accepting to take part in the experiment, participants were presented with the task instructions, which matched the demands of each condition. In the \textbf{Explanations} condition, after participants read the instructions, they went through an introduction on explanations and the interpretation of the local and global explanation plots. Then, they were presented with three multiple-choice questions testing whether they conceptually understood the distinction between local and global explanations and whether they were able to correctly interpret the explanation plots.  Participants in this condition needed to answer $2$ or $3$ questions correctly to be included in the sample.

Once the introduction was completed, participants in all conditions went through a familiarization phase, consisting of $12$ trials. In each trial, participants first inspected a person’s profile for whom the age, employer, education, marital status, occupation, ethnic background, gender and hours per week spent working were provided. Participants had to predict whether this person gains more or less than $100$K dollars per year and to give their confidence in their prediction by clicking on a Likert scale ranging from $1$ (low) to $7$ (high). In the next slide, participants were provided with the model’s assistance, which contained different kinds of information depending on the condition (see Section \ref{design1}) and they were asked to give their final prediction, which could be the same or different from their initial one. Once both of these steps were performed, participants were    shown the real life outcome for the person under consideration. The aim of our familiarization phase was two-fold. First, participants could understand better their task and develop some familiarity with the model's assistance (especially in the case of the \textbf{Explanations} condition, which contained a greater amount and a more diverse set of information) but more importantly, participants had the opportunity to gain some insight about the model’s performance. In particular, given that participants were provided with the real-life outcomes, they were exposed to instances where the model erred, from which they could infer that following the model blindly would not be a fruitful strategy. 

After the end of the familiarization phase, participants were informed that they were about to start with the main phase of the experiment, which consisted of $16$ trials. In each of them, the two first steps were identical to the first two steps of the familiarization phase, that is participants inspected an instance, they provided their prediction and their confidence in their prediction and in the next slide they were provided with the model’s assistance and they were required to provide their final prediction for this instance. In the test phase, however, after submitting their  final prediction, instead of inspecting the real life outcome, participants were asked to answer on  a scale from $1$ to $7$ to which extent they agreed with the following statements, which we borrowed from the scales in \citep{cahour2009does,adams2003trust}:

\begin{itemize}
    \item \textit{I relied on the model to reach my final prediction.}
    \item \textit{I understand the model's decision making process.}
    \item \textit{I trust the model's prediction for this person.}
\end{itemize}

Finally, after going through all $16$ trials, participants were presented with an exit survey of $9$ multiple choice questions which assessed their objective understanding of the model, adapted from \citep{wang2021explanations}.\footnote{All the question can be found in Appendix \ref{questionstrust}.} The aim of these questions was to address \textbf{H2}, since they allowed for comparing model understanding across conditions. This made possible to identify whether explanations offer any significant added benefits, compared to providing users with uncertainty estimates alone. The questions cover a wide spectrum of  objectives related to understanding:

\begin{itemize}
    \item \textbf{Global feature importance: } Participants were asked to select the most/least influential features the model utilizes to reach its predictions. ($2$ questions)
    \item \textbf{Local feature importance:} Participants were given a person's profile, and they were asked to select the most influential feature for this particular case. ($1$ question)
    \item \textbf{Counterfactual thinking:} Participants were presented with a person's profile, as well as a list of changes in the values of the features, and they were asked to select which of these changes would be sufficient to alter the model's prediction. ($2$ questions)
    \item \textbf{Model simulation:} Participants were given a profile, and they were asked to answer what they believed the model's prediction for this person would be. ($2$ questions)
    \item \textbf{Error detection: } Participants were shown a profile, as well as the model's prediction, and they were asked whether they find this prediction to be correct or not. ($2$ questions)
\end{itemize}

To make sure that participants were attentive, we included two attention checks in the experiment, where they  were given instructions about which answer they should submit. Those who failed to pass the checks, were excluded from the analysis. The base payment was $\$ 3.20$ for participants in the \textbf{Explanations} condition, and $\$ 3.00$ for the rest of them, since the former required participants to go through an introduction on feature importance explanations. Moreover, to further motivate participants, we included two performance based bonuses; those who provided a correct final prediction on more than $12$ of the $16$ main trials were given an extra $\$ 0.30$, and those who answered correctly more than $6$ of the questions in the exit survey received a bonus of $\$0.10$.

\section{Results}

\begin{figure}[h]

\centering
\subfloat[]{\label{initialfinal}
\centering
\includegraphics[scale=0.55]{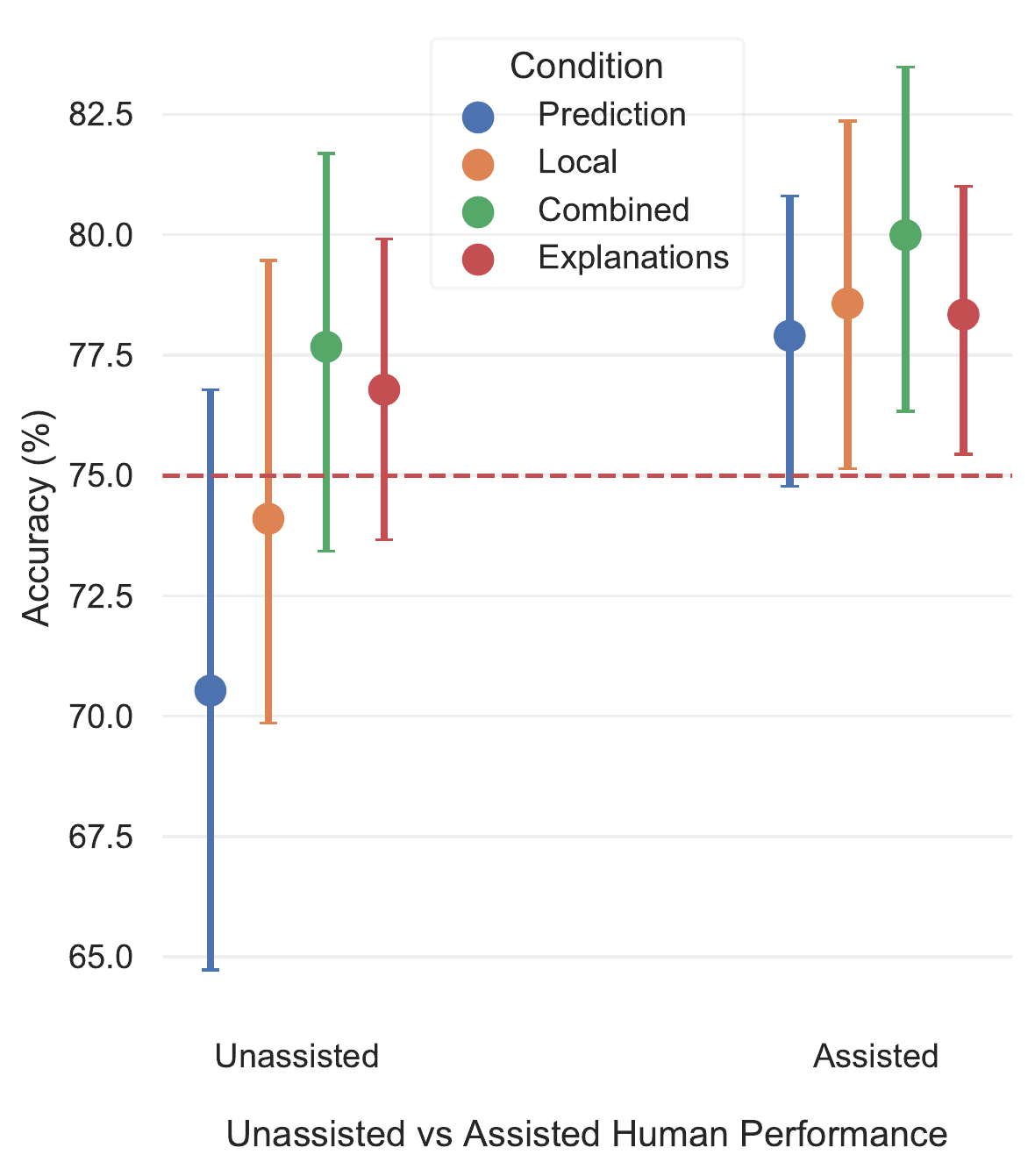}
}
\hfill
\subfloat[]{\label{conditionsdiff}
\centering
\includegraphics[scale=0.55]{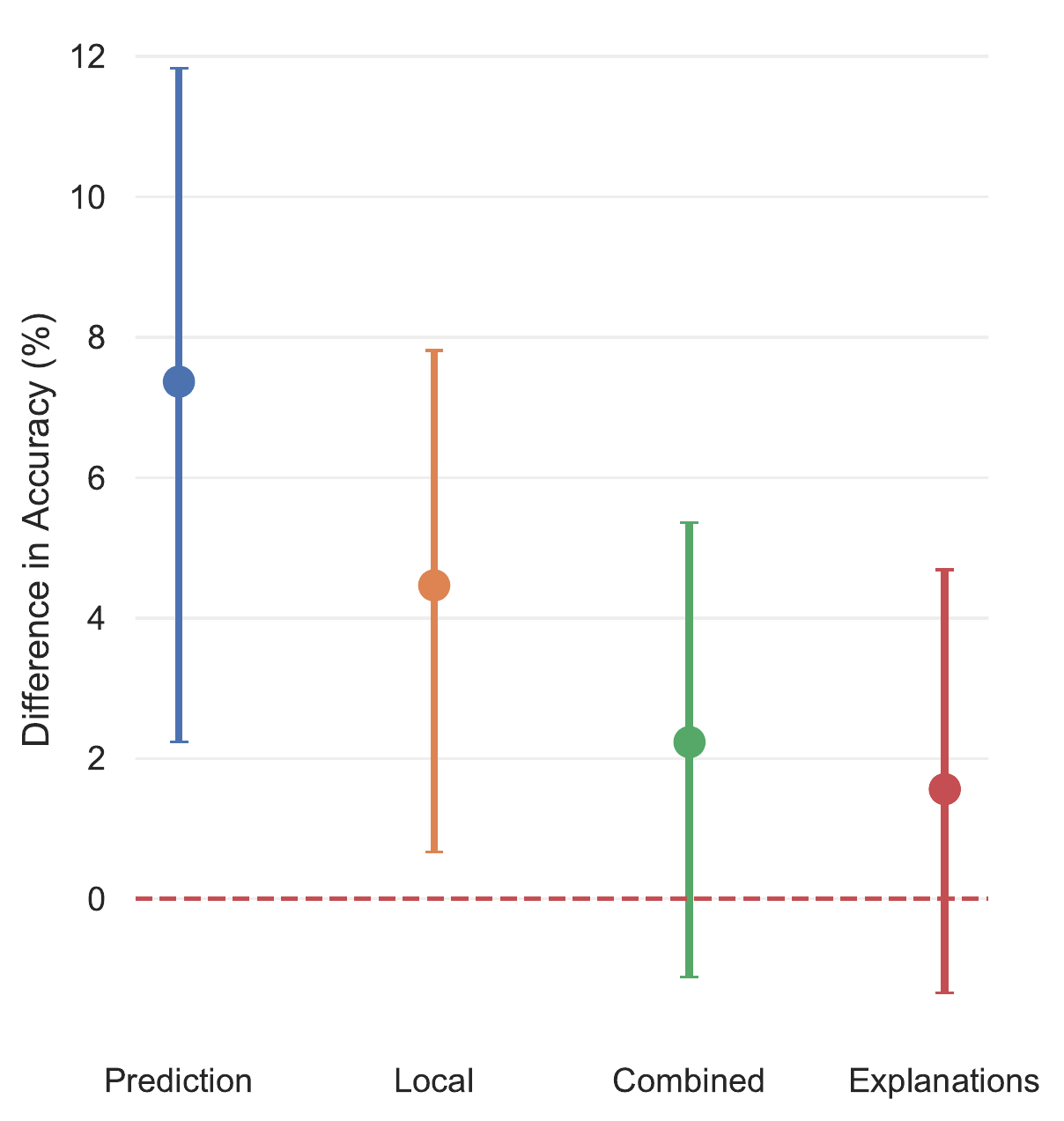}
}
\hfill
\subfloat[]{\label{modelaccuracy}
\centering
\includegraphics[scale=0.55]{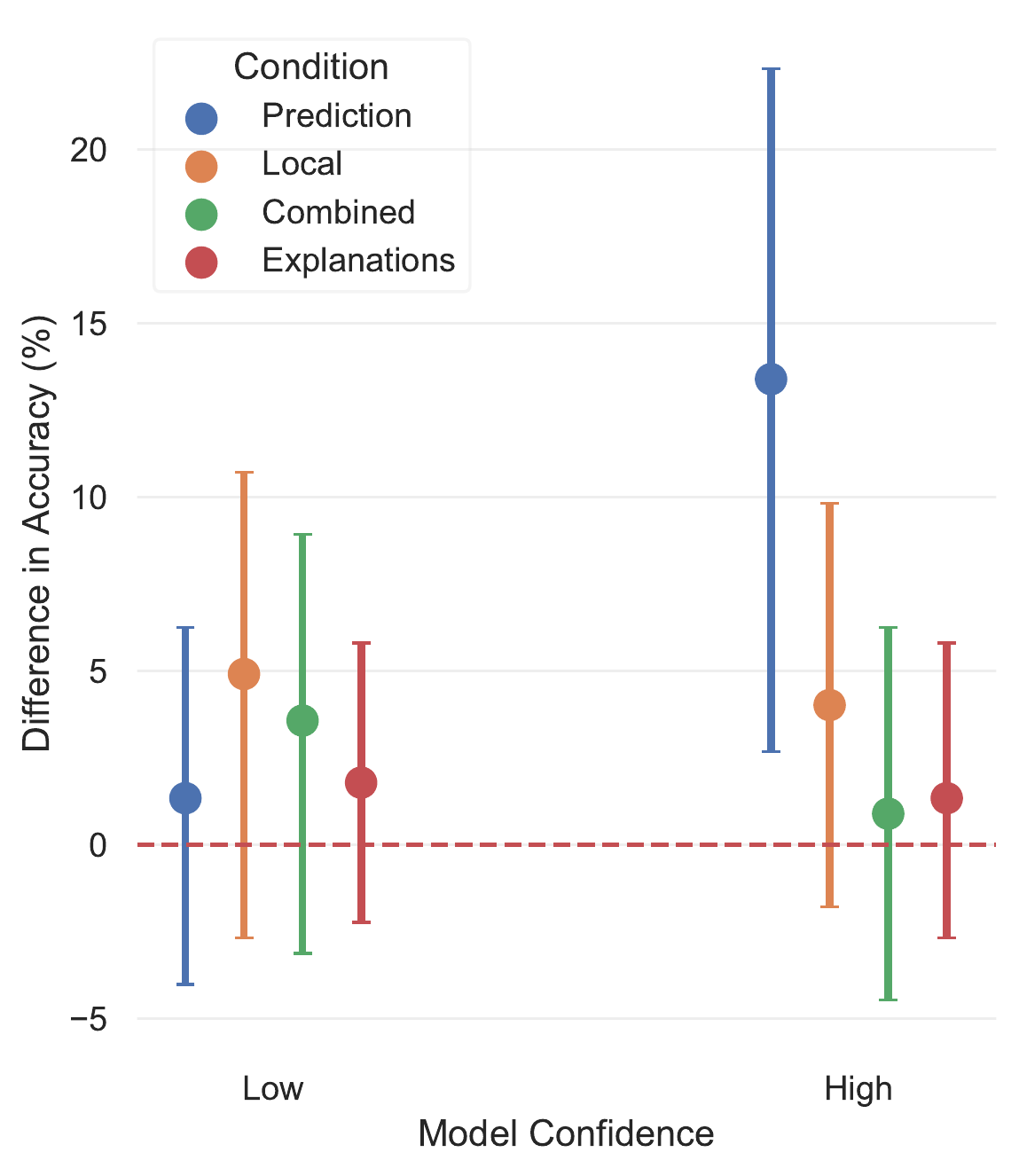}  
}
\caption{(a) Participants' unassisted and assisted accuracy. The red dotted line shows the model's accuracy. (b) Difference between participants assisted and unassisted accuracy, for each condition. (c) Difference in participants' accuracy as  a function of the model's confidence.}
\label{accuracies}
\end{figure}
In this section we present an analysis of our obtained data. All confidence intervals (CIs) were calculated using the non-parametric bootstrap estimation method \citep{efron1986bootstrap}. Pairwise comparisons between conditions were performed using the Mann-Whitney U Test \citep{mcknight2010mann}, while all other comparisons  were conducted using Wilcoxon's signed-rank test \citep{woolson2007wilcoxon}. Details about all CIs and comparisons can be found in Appendix \ref{effects}.

\subsection{Performance}\label{sec:perf}

The first set of analyses examined the effect of human confidence, model confidence, and model assistance (condition) on human performance. To address this question, we began with comparing the individual accuracy of the two parties, so that we can then assess whether the ensemble achieved superior performance. To this end, we compared participants’ accuracy before exposure to any model assistance (Unassisted Performance) to the model’s accuracy. Figure \ref{initialfinal}, depicts participants’ unassisted performance per condition, along with a $95\%$ confidence interval.  Details about all CIs are presented in Appendix \ref{effects}. Figure \ref{initialfinal} shows that $75\%$ belongs to all CIs, so participants and model showed comparable performance in all conditions, thus recreating the setting in \citep{bansal2021does}.


Then, we compared participants’ performance after exposure to the model's assistance (Assisted Performance) to the model’s accuracy. Figure \ref{initialfinal}, shows the assisted performance, along with the corresponding $95\%$ CIs.  Participants’ assisted performance was significantly better than $75\%$ in all but the \textbf{Prediction} condition, suggesting that even the simple strategy of pairing model predictions with confidence, as in the \textbf{Local} condition, is beneficial to participants’ performance, in line with the findings in \citep{bansal2021does}. On the other hand, participants in the \textbf{Prediction} condition failed to surpass the model’s performance, suggesting that predictions alone are not as effective in improving the joint performance, supporting the findings in \citep{lai2019human}.

Having established that the model's assistance helped the ensemble surpass the individual model accuracy, we continue by examining whether it surpassed participant's individual accuracy as well. Figure \ref{conditionsdiff}, shows the $95\%$ CIs of the difference between participants’ assisted and unassisted performance, per condition. Participants’ assisted performance was significantly better than their unassisted performance in the \textbf{Prediction} and \textbf{Local} conditions. On the contrary, the same comparison did not yield statistically significant results in the \textbf{Combined} and \textbf{Explanations} conditions, even though the point estimates were positive. This pattern can be explained, at least in part, by the fact that participants in the \textbf{Combined} and \textbf{Explanations} conditions already had better performance in their unassisted predictions compared to participants in the \textbf{Prediction} and \textbf{Local} conditions, leaving less room for improvement for them. Interestingly, when the point estimate of participants' unassisted accuracy was lower than the model's accuracy (conditions \textbf{Prediction} and \textbf{Local}), the ensemble surpassed the accuracy of both parties, however, when the point estimate was higher than $75\%$ (conditions \textbf{Combined} and \textbf{Explanations}), it failed to significantly outperform participants' individual accuracy. In \citep{bansal2021does}, participants' accuracy was always lower than the model's, so this might explain why the ensemble achieved superior accuracy in all tasks in their study. 

\begin{figure}
    \centering
    \includegraphics[scale=0.55]{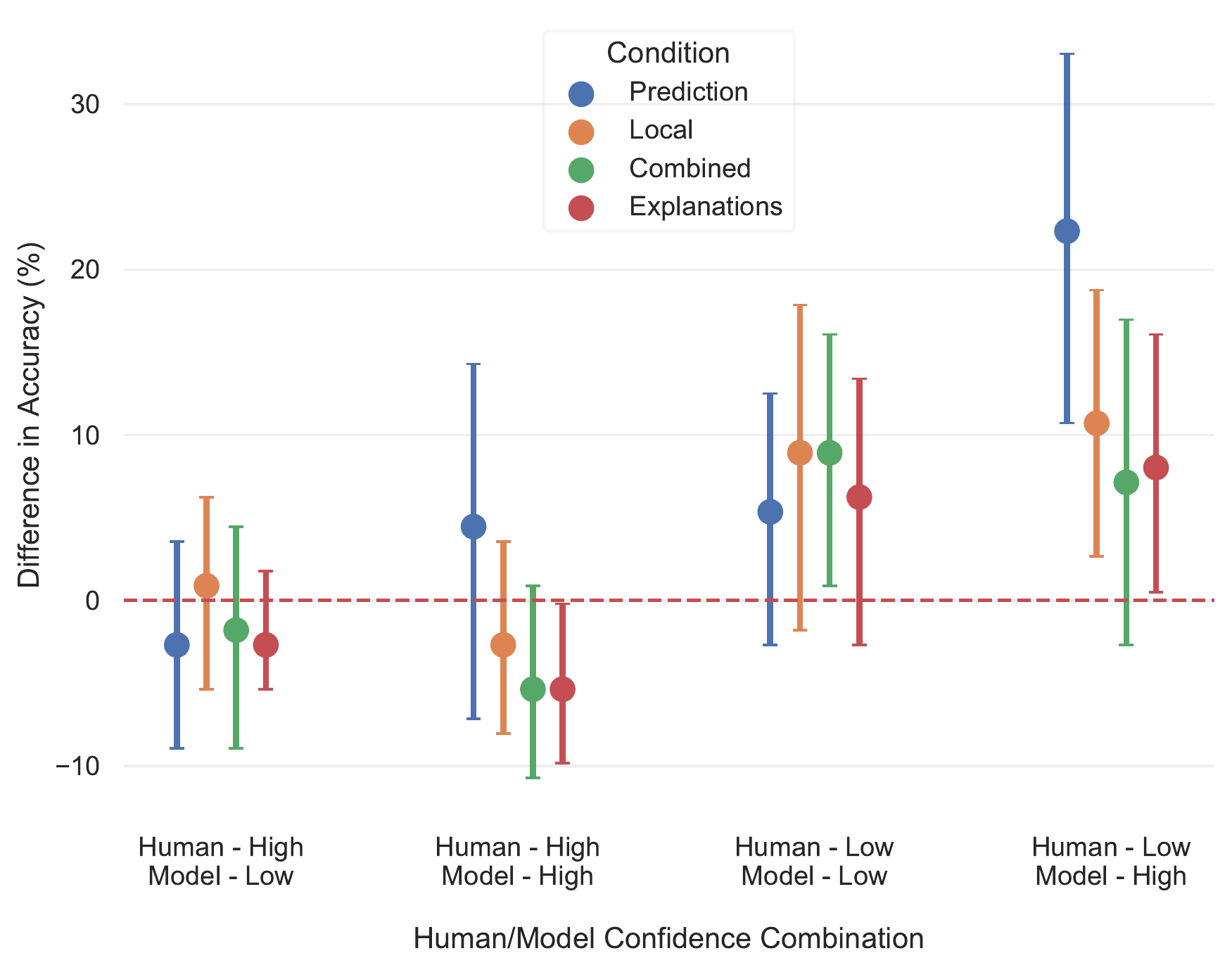}
    \caption{The difference between unassisted and assisted human performance, broken down by condition, human confidence, and model confidence. The red line shows the model's accuracy.}
    \label{humanmodelacc}
\end{figure}

Expanding on the above findings, we then isolated the effect of the different levels of model confidence (Low/High) on participants' accuracy (see Figure \ref{modelaccuracy}).  The resulting analysis showed that, with the exception of the \textbf{Prediction} condition, model confidence did not appear to modulate participants’ performance. Note that the \textbf{Prediction} condition was the only one where participants had in fact no information about whether the model had low or high confidence, and taking into account the width of the corresponding CI, which suggests that there was substantial variation in participants' accuracy, this result might be due to noise in the data.


A careful inspection of the pattern of the results discussed so far leads to a seemingly paradoxical observation: Focusing on the \textbf{Local} condition, we found that the model's assistance significantly improved participants' performance, yet when we broke down this effect for the different levels of model confidence, neither high nor low confidence model predictions significantly improved participants' performance. This leads to the puzzling conclusion that when considering model assistance in general, it helped participants improve their accuracy, but  when zooming into each of the levels of assistance it provides (low and high confidence predictions) separately, this effect vanishes. This phenomenon is known as the \textit{Simpson's paradox}, and it has been extensively studied in statistics, causal inference and philosophy \citep{wagner1982simpson,julious1994confounding,hernan2011simpson}. In statistical terms, this indicates that there are important confounding variables and/or causal relationships, that have not been accounted for into the analysis. The emergence of this phenomenon in our analysis perfectly captured the potential perils of not taking into account human confidence, since as soon as we adjusted for this factor, the  paradox resolved itself.

Figure \ref{humanmodelacc} breaks down the difference between assisted and unassisted accuracy, as a function of condition, human confidence, and model confidence. Participants’ accuracy showed a significant improvement when they were themselves uncertain, but the model showed high confidence in its predictions, in all but the \textbf{Combined} condition (see Appendix \ref{effects}), suggesting that the significant effect observed in the \textbf{Local} condition (discussed in the previous paragraph) was due to   this interaction, which is why looking at model confidence alone resulted into seemingly paradoxical conclusions. Furthermore, for the \textbf{Combined} condition, we found a significant improvement when both model and human confidence were low. These findings demonstrated that although we found no significant overall improvement for participants in the \textbf{Combined} and \textbf{Explanations} conditions, interpreting our results through the interaction of human and model confidence allowed us to detect fine grained effects that would have been otherwise missed.



\begin{figure}[h]

\centering
\subfloat[]{\label{humanmodelreliance}
\centering
\includegraphics[scale=0.45]{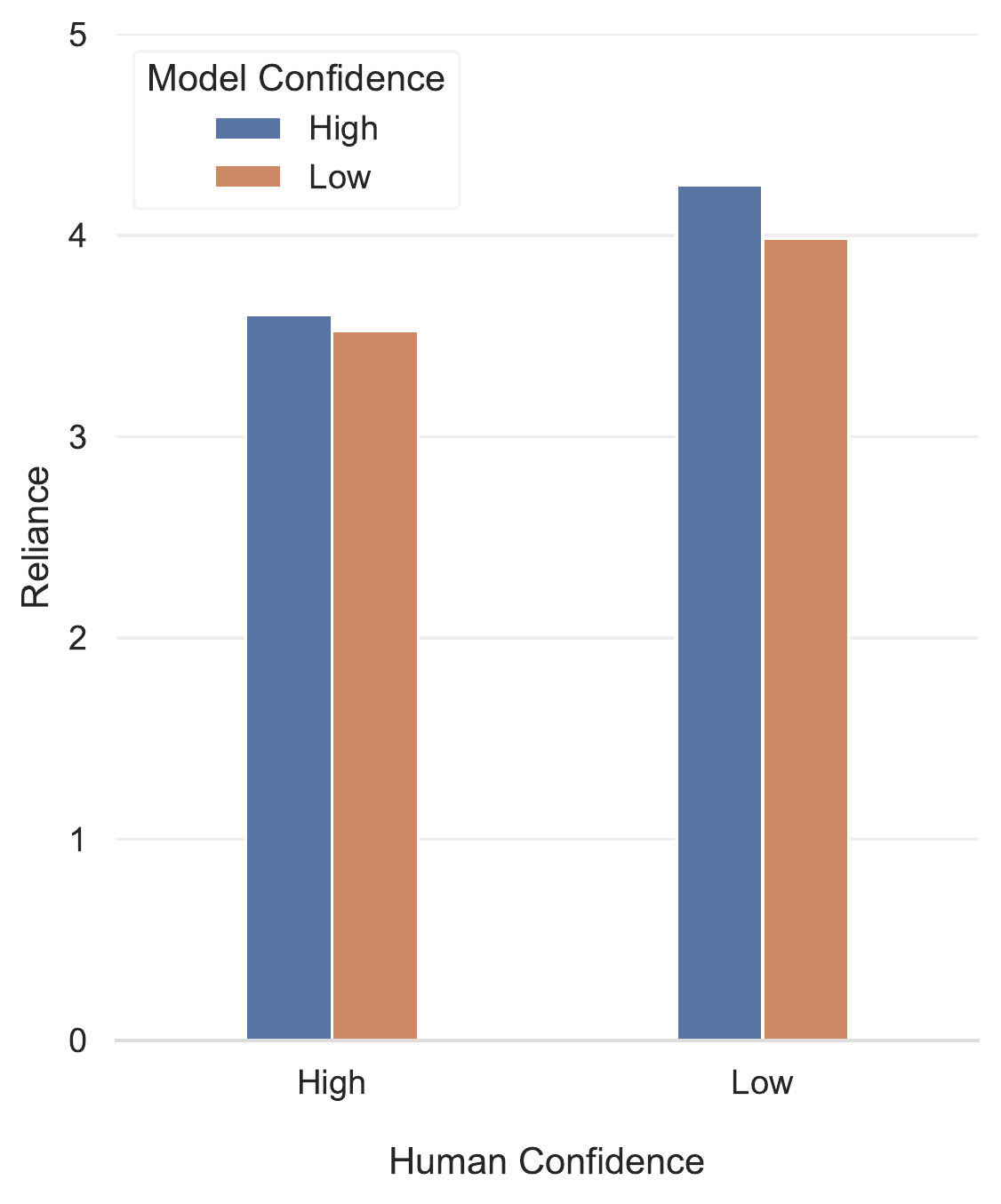}  
}
\hfill
\subfloat[]{\label{condmodelrel}
\centering
\includegraphics[scale=0.45]{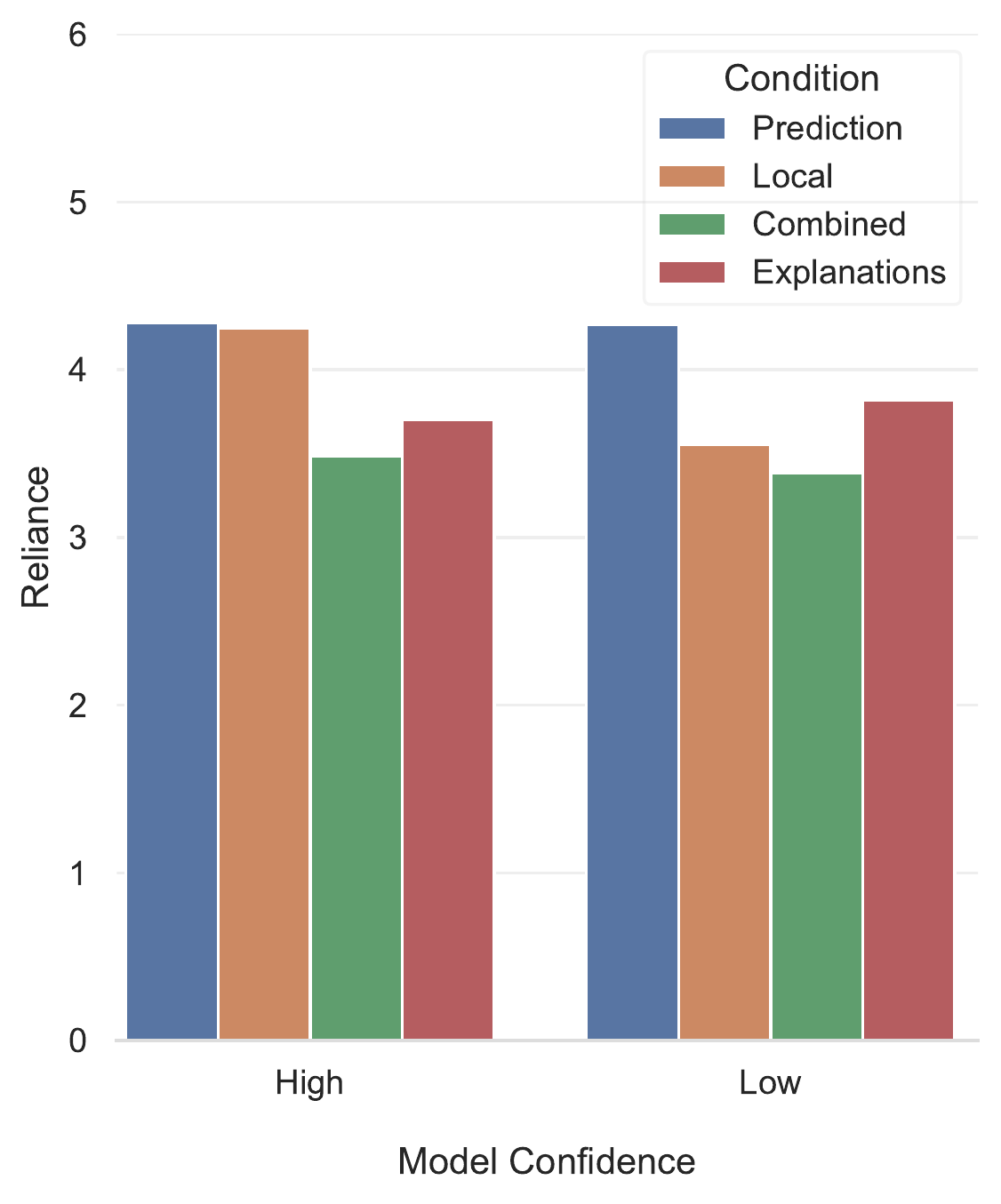}  
}
\caption{(a) Differences in reliance with respect to the interaction of human and model confidence. (b) Differences in reliance with respect to the interaction of condition and model confidence.}
\label{reliances}
\end{figure}

On the other hand, when participants were confident about their predictions, but the model was not, there was virtually no difference in accuracy, indicating that participants' predictions were primarily driven by their own intuitions or knowledge of the world. Finally, when both parties were confident in their predictions, participants’ performance slightly declined, but this effect only reached significance in the \textbf{Explanations} condition. A possible interpretation of this pattern is that explanations and high model confidence prompted participants to exhibit a slightly over-reliance on the model, which is consistent with the findings in \citep{kaur2020interpreting}. The fact that the reverse trend was observed in the \textbf{Prediction} condition strengthens this interpretation, suggesting that in the  absence of uncertainty estimates, participants' own confidence dominated, thus no over-reliance was observed. These findings provide strong evidence in favour of \textbf{H1}, suggesting that the interaction between human and model confidence is an important factor influencing  when and how much  a model's predictions will be followed, above and beyond   model confidence.

\subsection{Reliance, Understanding, and Trust}

This set of analyses examines the effect of human confidence, model confidence and condition on participants reliance, understanding, and trust. Following the discussion in \citep{wobbrock2016nonparametric}, we opted for analyzing our data using a semi-parametric ANOVA approach, which is robust against violations of the underlying parametric  ANOVA assumptions, such as normality, in line with numerous recent studies \citep{roo2017one,gugenheimer2017sharevr,hartmann2019realitycheck,thoravi2020transceivr,kudo2021towards} that utilize non- or semi-parametric methods. In particular, we based our analysis on the Wald-type statistic  proposed in \citep{konietschke2015parametric}.



\subsubsection{Reliance}\label{sec:rel}



Starting with reliance, a three-way repeated measures ANOVA with Human Confidence$\times$Model Confidence$\times$Condition identified a main effect of Human Confidence ($W(1)=40.17$, $p<0.001$), a main effect of Model Confidence ($W(1)=5.138, p=0.023$), as well as an interaction between Condition and Model Confidence ($W(3) = 17.574, p=0.001$). Participants’ reliance dropped by $7.8\%$ when they themselves were confident, compared to when they were uncertain. Moreover, participants' reliance increased by $2.4\%$ when the model made high confidence predictions. Contrasting these two effect sizes, we see that the former is more than $3$ times bigger than the latter, providing evidence that it is primarily human confidence that influences model reliance, in line with \textbf{H3.1}. However, overall this hypothesis was only partially confirmed, since we did not detect a significant interaction between  human and model confidence ($W(1)= 1.344, p=0.246$). That being said,  we suspect this was due to sample size limitations, since the combination of confidences (Human - Low \& Model - High) showed the greatest reliance, suggesting that a bigger sample size would lead to statistically significant results (see Figure \ref{humanmodelreliance}). 



\begin{figure}[h]

\centering

\subfloat[]{\label{humanmodelunder}
\centering
\includegraphics[scale=0.45]{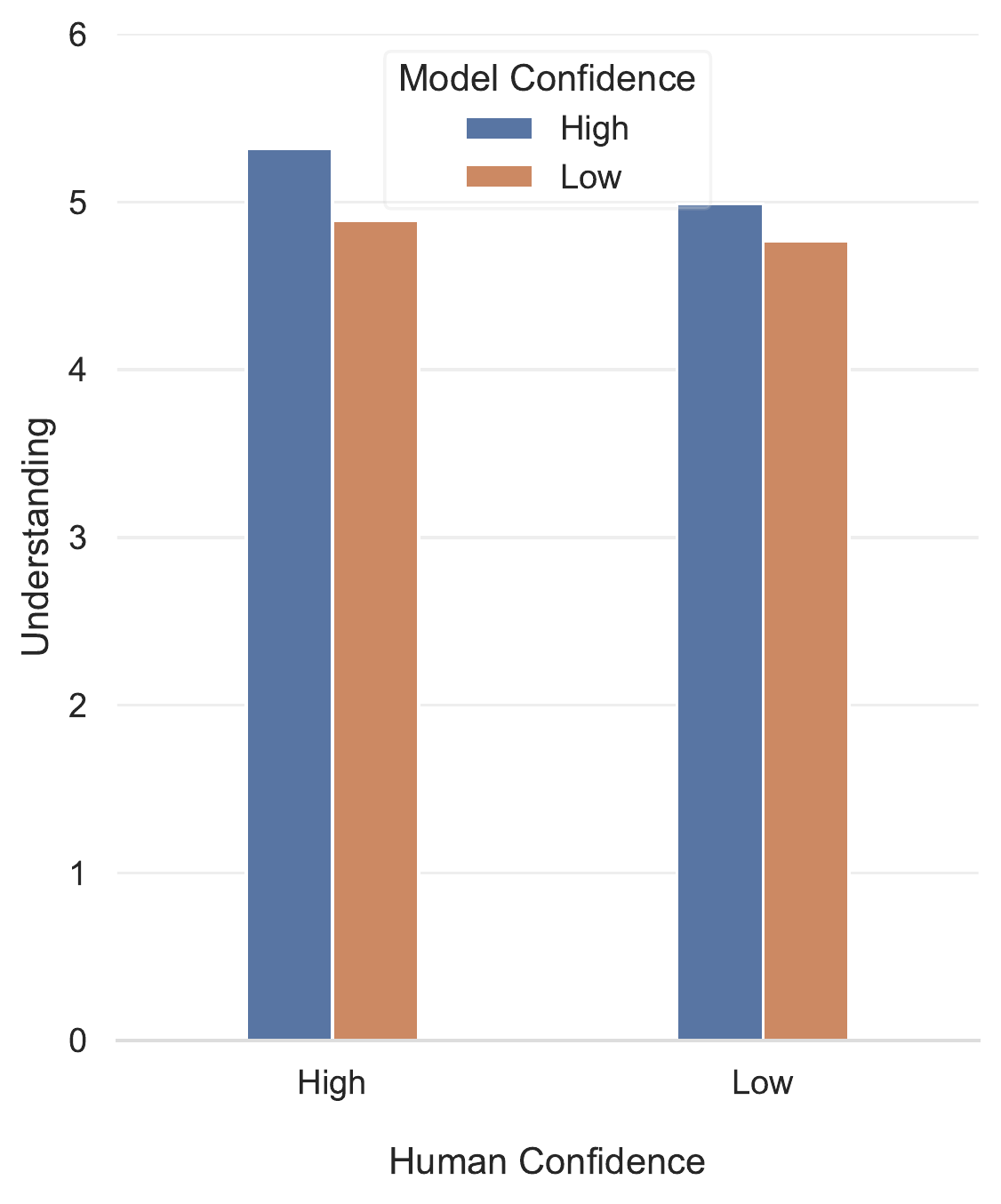}  
}
\hfill
\subfloat[]{\label{condunder}
\centering
\includegraphics[scale=0.45]{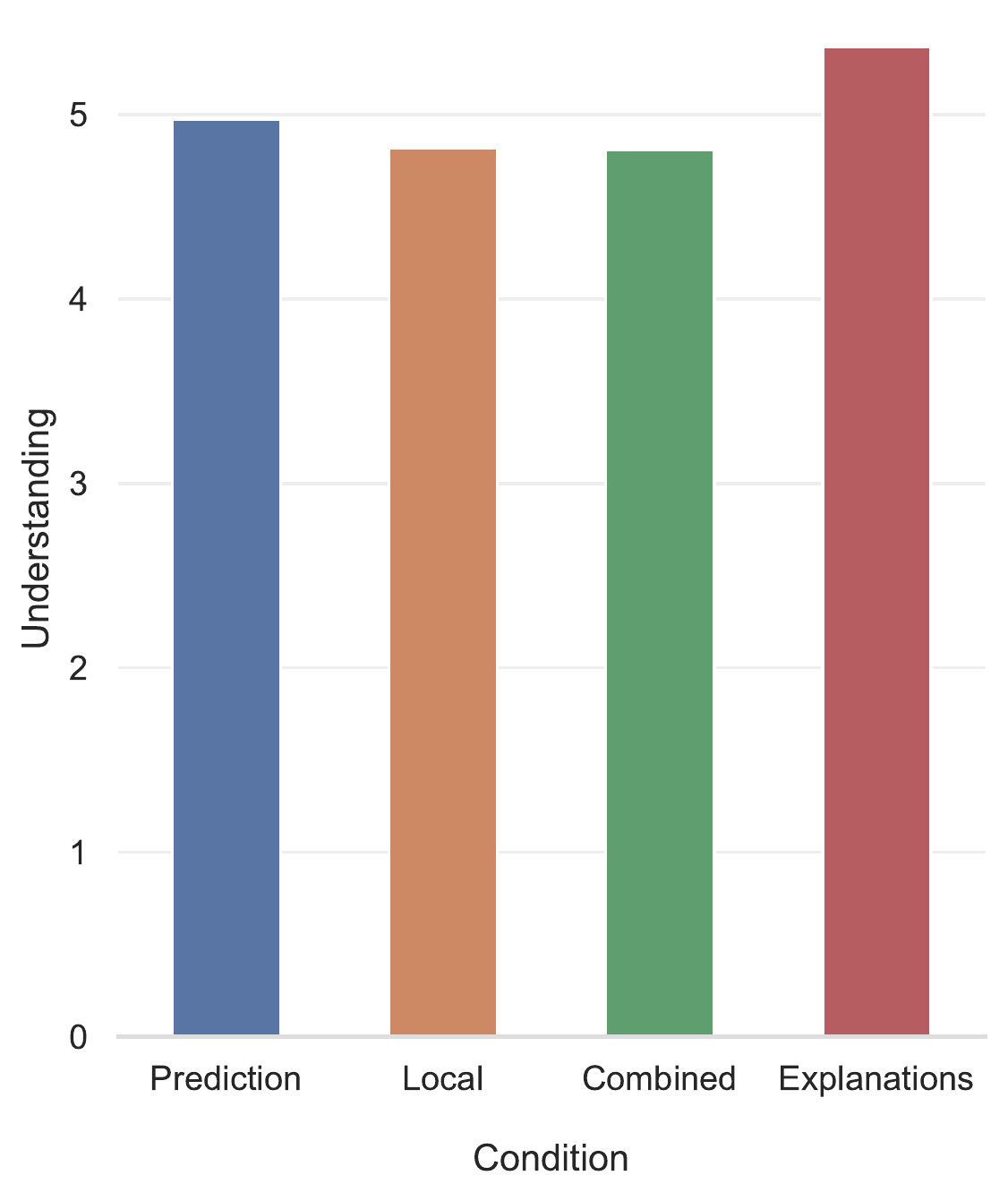}  
}
\caption{(a) Differences in understanding with respect to the interaction of human and model confidence. (b) Differences in understanding with respect to each condition. }
\label{accuracies}
\end{figure}

With respect to the interaction between Condition and Model Confidence, pairwise comparisons revealed that this effect was due to the \textbf{Local} condition ($Z=32, p<0.001$). Moreover, as Figure \ref{condmodelrel} shows the remaining conditions exhibited virtually no variation in reliance for the different levels of model confidence. In the \textbf{Local} condition, participants' reliance was $9.8\%$ higher when the model was confident, compared to when it was not. A possible interpretation of this finding is that while local confidence communicates model uncertainty, it does not provide any meta-information quantifying the robustness of this information, thus it did not allow participants to adjust their reliance. This is because they were only aware of the model's uncertainty, but they did not have any information about either the model's global error rates (as in the \textbf{Combined} condition) or about the reasons behind the prediction (as in the \textbf{Explanations} condition). This is a very interesting finding that demonstrates that although extra information might not necessarily lead to better predictive accuracy, it can play a major part in adjusting human behaviour.

\subsubsection{Understanding} \label{sec:understannding1}
Moving on we turn our attention to participants understanding, and how it was impacted by the various factors in our study. A three-way repeated measures  ANOVA with Human Confidence$\times$Model Confidence$\times$Condition identified a main effect of Human Confidence ($W(1)=18.114, p<0.001$), a main effect of Model Confidence ($W(1)=23.015, p<0.001$), a main effect of Condition ($W(3)=10.944, p=0.012$), as well as an interaction between Human Confidence and Model Confidence ($W(1)=3.963, p=0.047$). Participants’ subjective understanding improved by $3.2\%$, when they had high confidence, suggesting that they took into account their own knowledge when interpreting the model’s predictions. Moreover, participants' understanding improved by $4.6\%$ when the model was confident, compared to when it was not, providing evidence that high confidence model predictions made participants feel more certain that their understanding was correct. With respect to the interaction of human and model confidence, pairwise comparisons revealed that when both human and model confidence were high, understanding was significantly higher than all the remaining combinations. In more detail, compared to the combinations (Human - High \& Model - Low), (Human - Low \& Model - High), (Human - Low \& Model - Low), understanding was $4.72\%$ ($Z=1755, p<0.001$), $6.41\%$ ($Z=1555, p<0.001$), and $7.84\%$ ($Z=1148.5, p<0.001$), higher, respectively. This provided evidence that the interaction of human and model confidence influences model understanding, which fully supported \textbf{H3.2}. No other comparison yielded significant differences (see Figure \ref{humanmodelunder} and Appendix \ref{effects})







\begin{figure}[h]

\centering

\subfloat[]{\label{humanmodeltrust}
\centering
\includegraphics[scale=0.45]{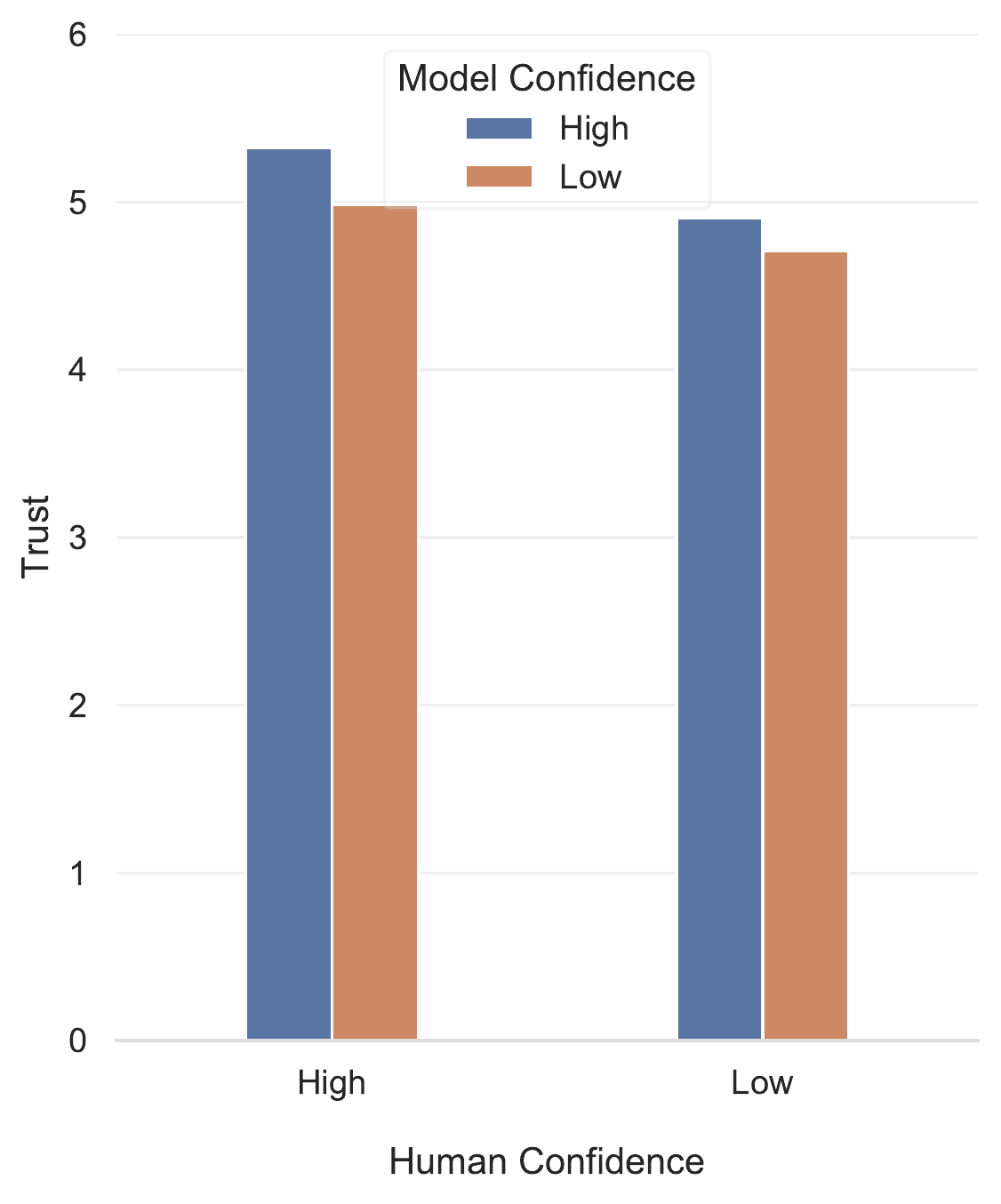}  
}
\hfill
\subfloat[]{\label{condmodeltrust}
\centering
\includegraphics[scale=0.45]{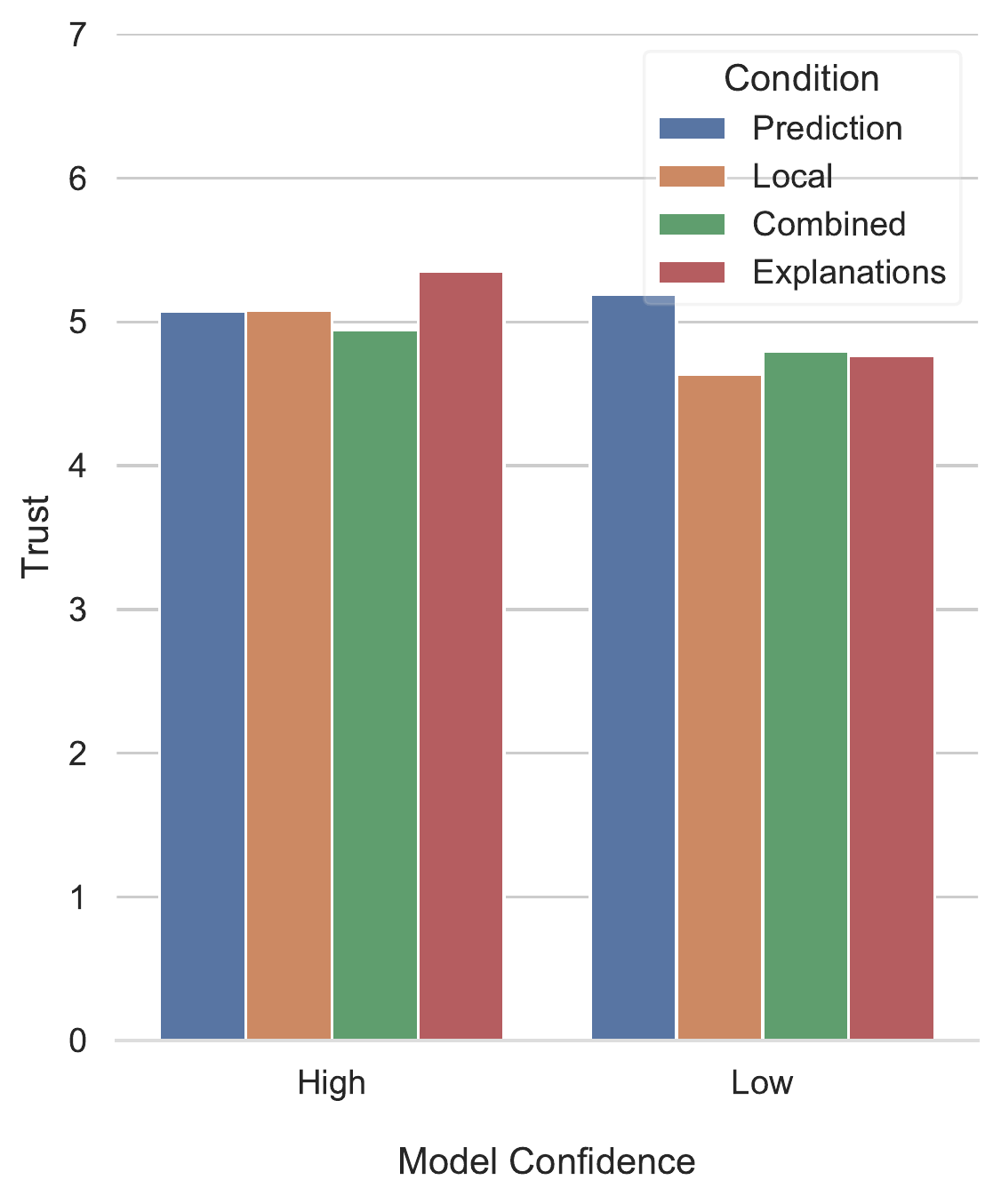}  
}
\caption{(a) Differences in trust with respect to the interaction of human and model confidence. (b) Differences in trust with respect to the interaction of condition and model confidence. }
\label{accuracies}
\end{figure}

Finally, looking at the main effect of Condition, pairwise comparisons showed that subjective understanding ratings in the \textbf{Explanations} condition differed significantly from the ones in the \textbf{Local} ($U=-2.5, p=0.0365$) and \textbf{Combined} ($U=-3.01, p=0.007$) conditions, but not from the ones in the \textbf{Prediction} condition ($U=-1.13, p=0.774$). Figure \ref{condunder} shows the average subjective understanding per condition. The fact that there was no difference between the \textbf{Explanations} and \textbf{Prediction} conditions, is consistent with the finding that humans tend to project their reasoning on the model, without actually having a well-versed understanding of the model's decision making process. In contrast, in the \textbf{Local} and \textbf{Combined} conditions, participants were aware of the model's uncertainty, so they were more conservative with their understanding scores. The actual discrepancy of model understanding between the \textbf{Explanations} and \textbf{Prediction} conditions will become more apparent in Section \ref{objectiveunder}, where we discuss participants' objective model understanding.

\subsubsection{Trust}\label{sec:trust}

We concluded this part of the analysis studying participants' trust towards the model's predictions. A three-way repeated measures  ANOVA with Human Confidence$\times$Model Confidence$\times$Condition identified a main effect of Human Confidence ($W(1)=46.269, p<0.001$), a main effect of Model Confidence ($W(1)=12.942, p<0.001$), as well as an interaction between Condition and Model Confidence ($W(3)=14.817, p=0.002$). Participants’ trust increased by $5\%$ when they were confident in their predictions. Moreover, participants' trust increased by $3.7\%$ when  model confidence was high. The difference in size between these two effects suggests that while both influenced participants' trust, the uncertainty stemming due to their own confidence had a slightly more pronounced effect. Despite the fact that we did not find significant evidence in favour of the effect arising from the interaction between human and model confidence ($W(1)=1.358, p=0.244$), we suspect that this was mainly due to sample size limitations, since the pattern shown in Figure \ref{humanmodeltrust}, suggested that when both parties were confident, participants' trust was likely higher. As it was the case when studying reliance, \textbf{H3.3} was partially supported,  calling for further investigations on the effect of the interaction of human and model confidence on trust.

Finally, following up on the interaction between Condition and Model Confidence, pairwise comparisons revealed that  in the \textbf{Local} ($Z=88, p=0.035$) and \textbf{Explanations} ($Z=77, p=0.016$) conditions participants tended to trust high confidence model predictions more than low ones (see Figure \ref{condmodeltrust}). In the \textbf{Local} condition, high confidence model predictions improved trust ratings by $6.3\%$. In the \textbf{Explanations} condition, this difference was even more pronounced, and equal to $8.4\%$. There is a rather intuitive interpretation of this result, in the sense that when participants were presented with local confidence information, it was reasonable that high confidence predictions imparted higher levels of trust. However, when these scores were complemented with global error rates, participants became aware of the fact that high confidence predictions might not necessarily translate into high accuracy, which is why they did not induce the same level of trust ($Z=160, p=1$). Having said that, when all this information was paired with explanations, participants were able to inspect the model's reasoning for each individual instance, so high confident predictions paired with reasonable explanations bypassed the uncertainty induced due to poor global error rates (as when the model predicts More than $100$K dollars).

\begin{figure}
    \centering
    \includegraphics[scale=0.55]{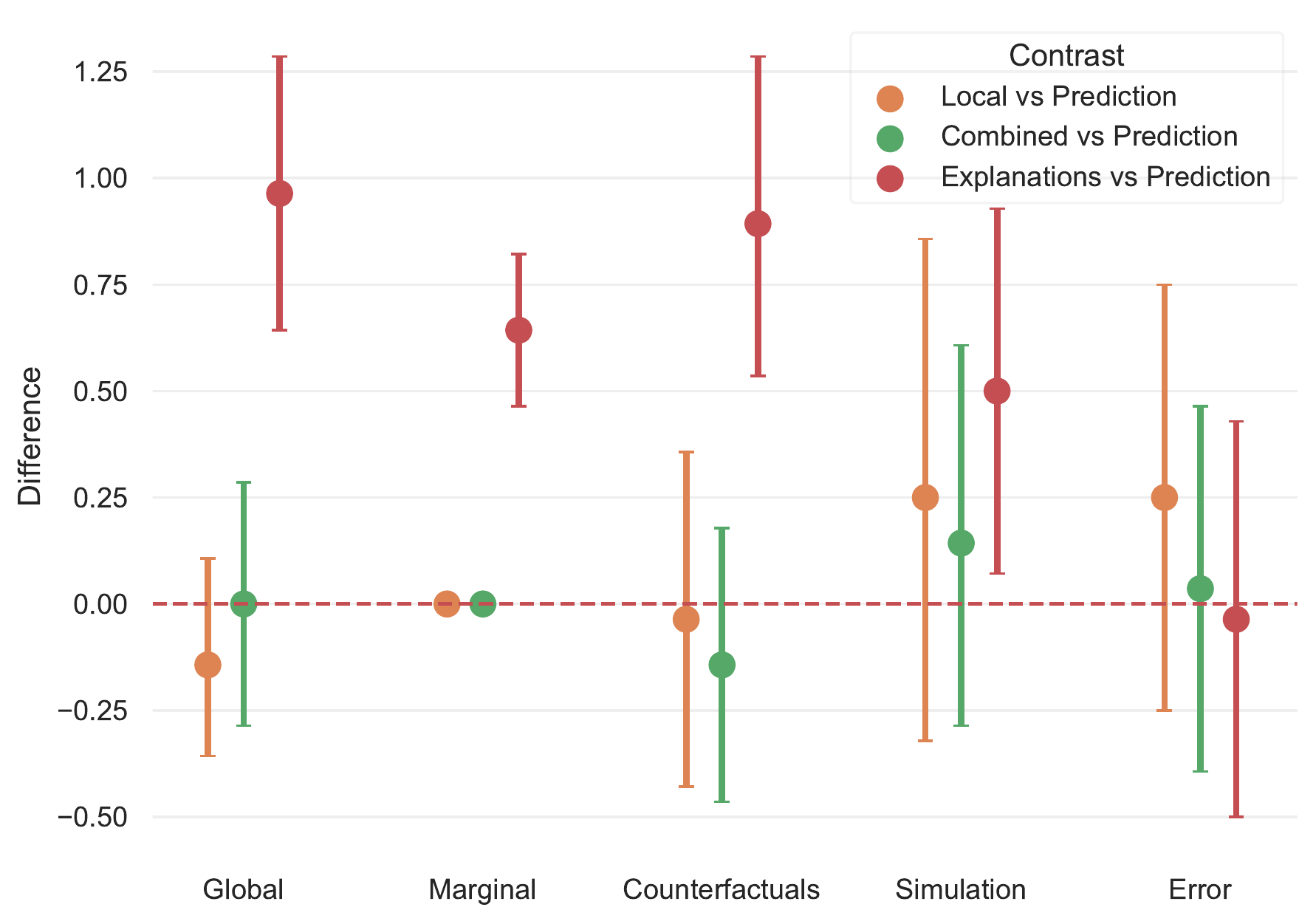}
    \caption{Difference between \textbf{Prediction} and every other condition, for each aspect of model understanding.}
    \label{testobjective}
\end{figure}

\subsection{Objective Understanding}\label{objectiveunder}

In this section we studied objective model understanding, as captured via the $9$ multiple choice questions that participants completed before exiting the experiment. We looked for differences between \textbf{Prediction} and every other condition,  to assess whether including uncertainty estimates or explanations led to improved understanding, compared to providing model predictions alone. Recall that these questions addressed $5$ different aspects of objective model understanding. Each aspect is analyzed separately in order to gain a more refined picture of participants' understanding. Figure \ref{testobjective}, shows the difference in scores between conditions, broken down by each aspect of understanding. Starting with global feature importance, participants’ scores in the \textbf{Explanations} condition significantly outperformed those in the \textbf{Prediction} one, while there was no difference between the remaining contrasts. This result was not surprising since global feature importance information was available to participants in the \textbf{Explanations} condition. However, the fact that there was  no difference among the remaining conditions highlighted that uncertainty estimates were as effective as plain predictions in helping participants infer such information.

With respect to local feature importance the discrepancy was even more severe, since no participant in the \textbf{Prediction}, \textbf{Local}, \textbf{Combined} conditions was able to provide a correct answer. On the other hand, $64.3\%$ of the participants in the \textbf{Explanations} condition answered this question correctly. Again, we expected participants in the latter to have an edge on this task, however, in contrast to global feature importance which remains constant across instances, local feature importance information depends on the instance at hand, meaning that this effect was not due to mere memorization. Instead, participants needed to critically reflect on the information presented throughout the experiment to reach their decision. This sharp difference clearly demonstrated that when it came to inferring local feature importance the information in the remaining conditions was insufficient.

Participants’ scores in the counterfactual component of the test showed again that only those in the \textbf{Explanations} condition significantly outperformed those in the \textbf{Prediction} condition. This is a very interesting finding,  indicating that although explanations contained factual information, participants were able to extract counterfactual knowledge out of them, while uncertainty information did not provide any such benefits. The exact same pattern was observed when considering the aspect of model simulation, despite the fact that explanations themselves did not explicitly contain any information regarding simulating the model's behaviour. Regardless, the enhanced understanding of the model's decision making process helped participants in the \textbf{Explanations} condition achieve superior performance in the simulation component of the test.


Finally, participants’ ability to detect erroneous model predictions was assessed, where no significant differences between conditions were found. Error detection closely resembled the main prediction task, since it required inspecting an instance and the corresponding model prediction to assess its correctness. This means participants in all conditions had substantial exposure/familiarity with this procedure, which explains why there was no difference in their performance. Overall, the preceding analysis provided strong evidence suggesting that explanations led to better model understanding, compared to uncertainty estimates, thus fully supporting \textbf{H2}.

\begin{figure}[]

\centering
\subfloat[Human - High \& Model - High]{\label{highhighswitch}
\centering
\includegraphics[scale=0.42]{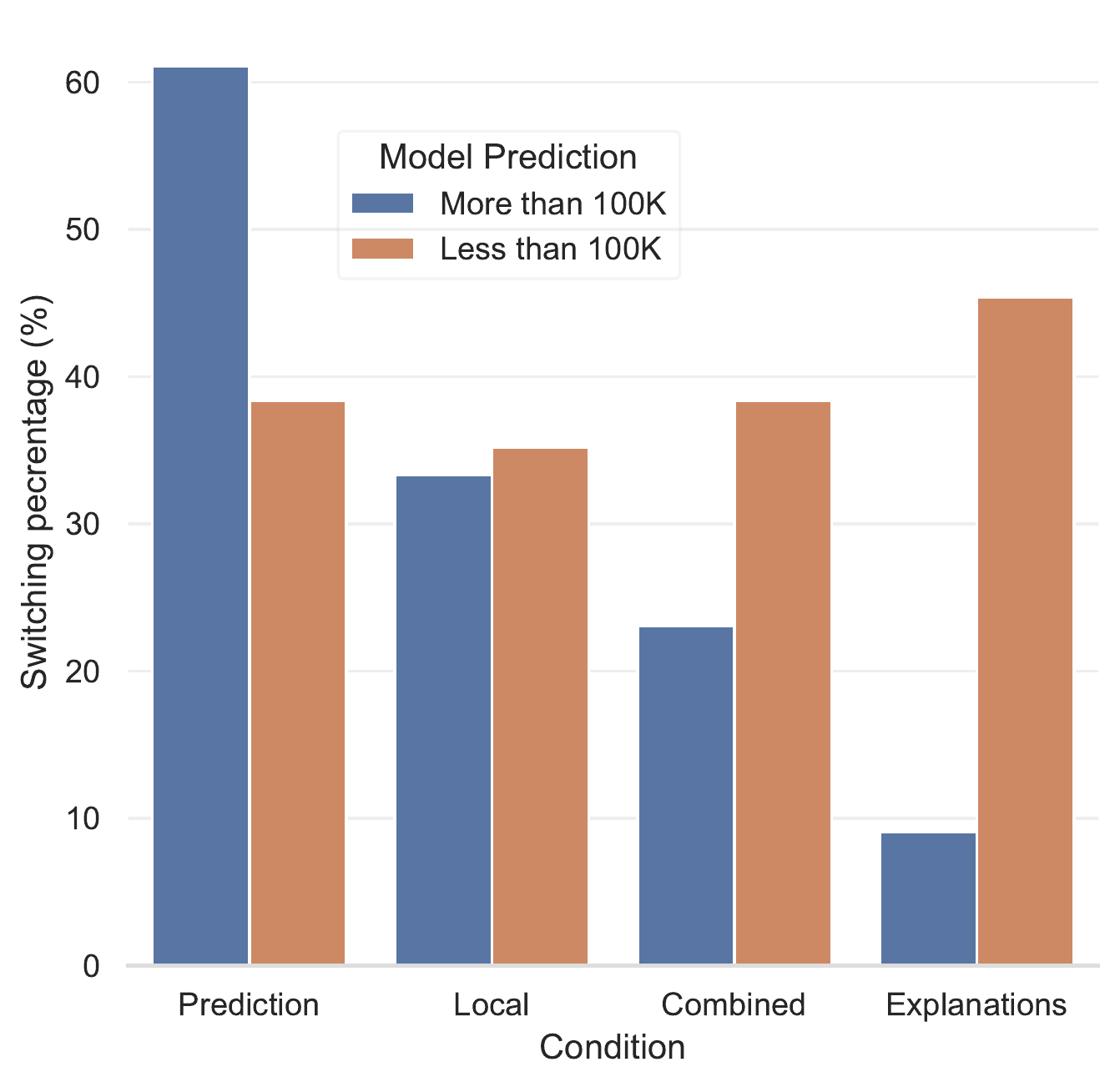}
}
\hfill
\subfloat[Human - High \& Model - Low]{\label{highlowswitch}
\centering
\includegraphics[scale=0.42]{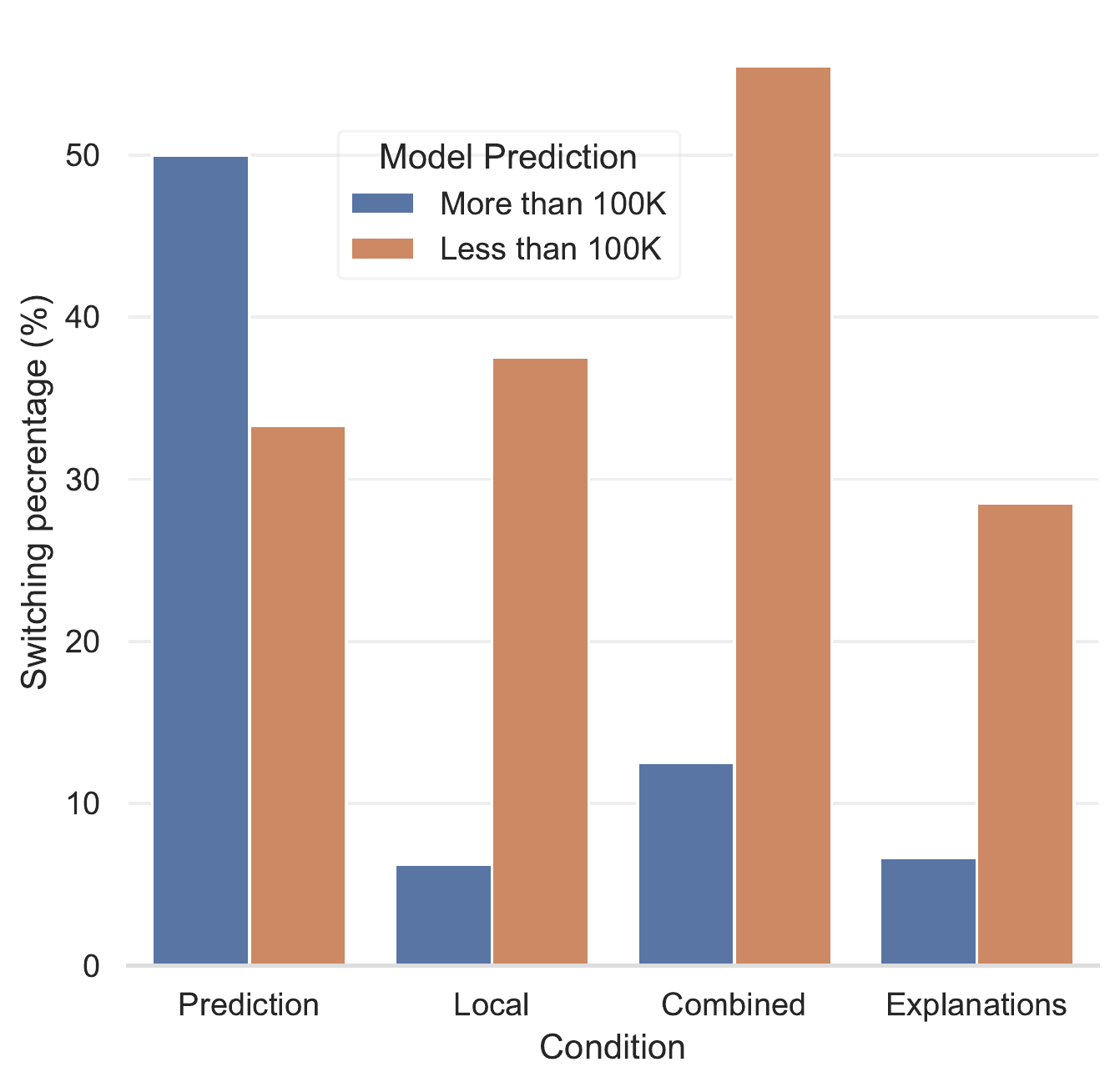}
}

\subfloat[Human - Low \& Model - High]{\label{lowhighswitch}
\centering
\includegraphics[scale=0.42]{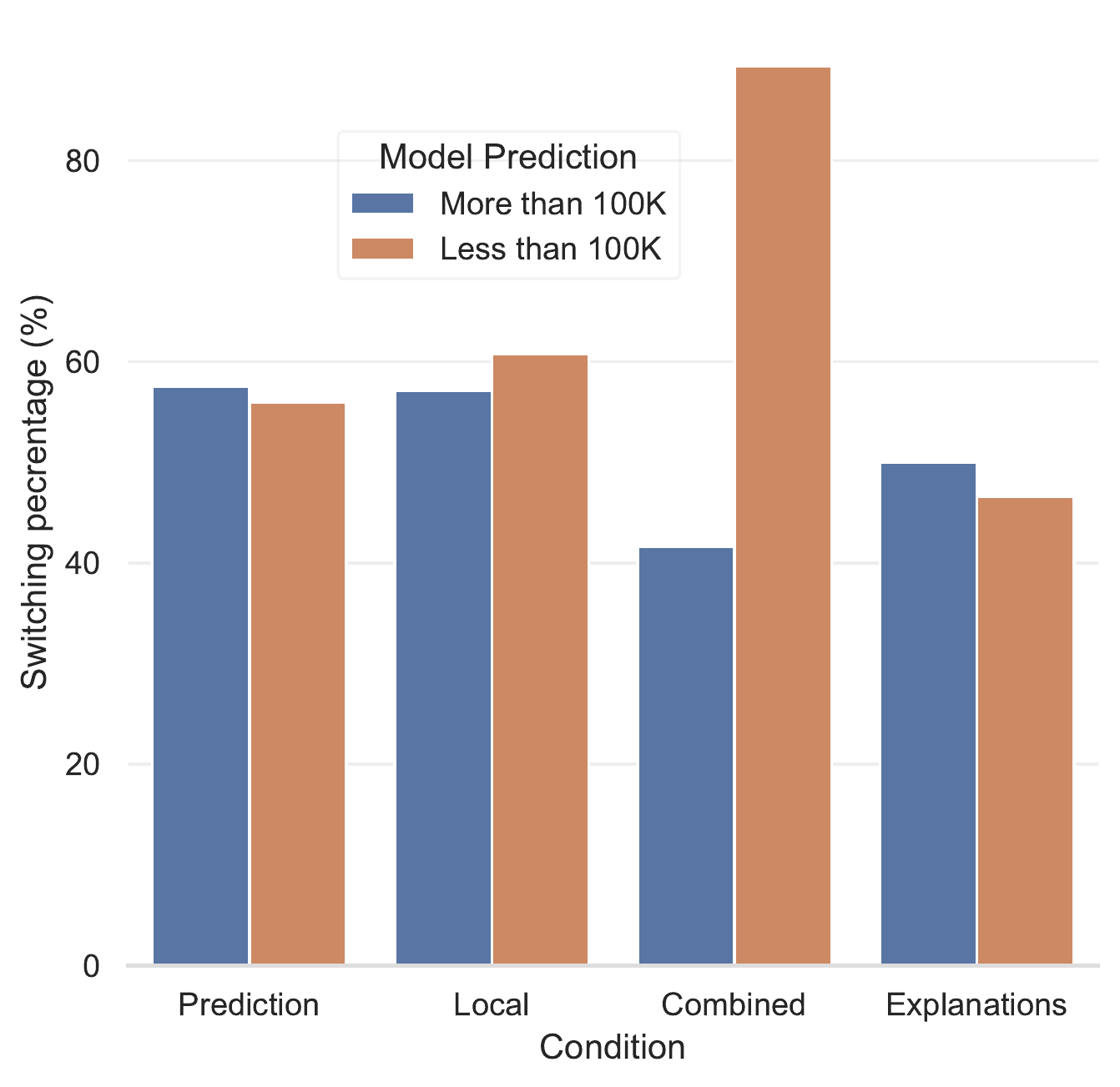}  
}
\hfill
\subfloat[Human - Low \& Model - Low]{\label{lowlowswitch}
\centering
\includegraphics[scale=0.42]{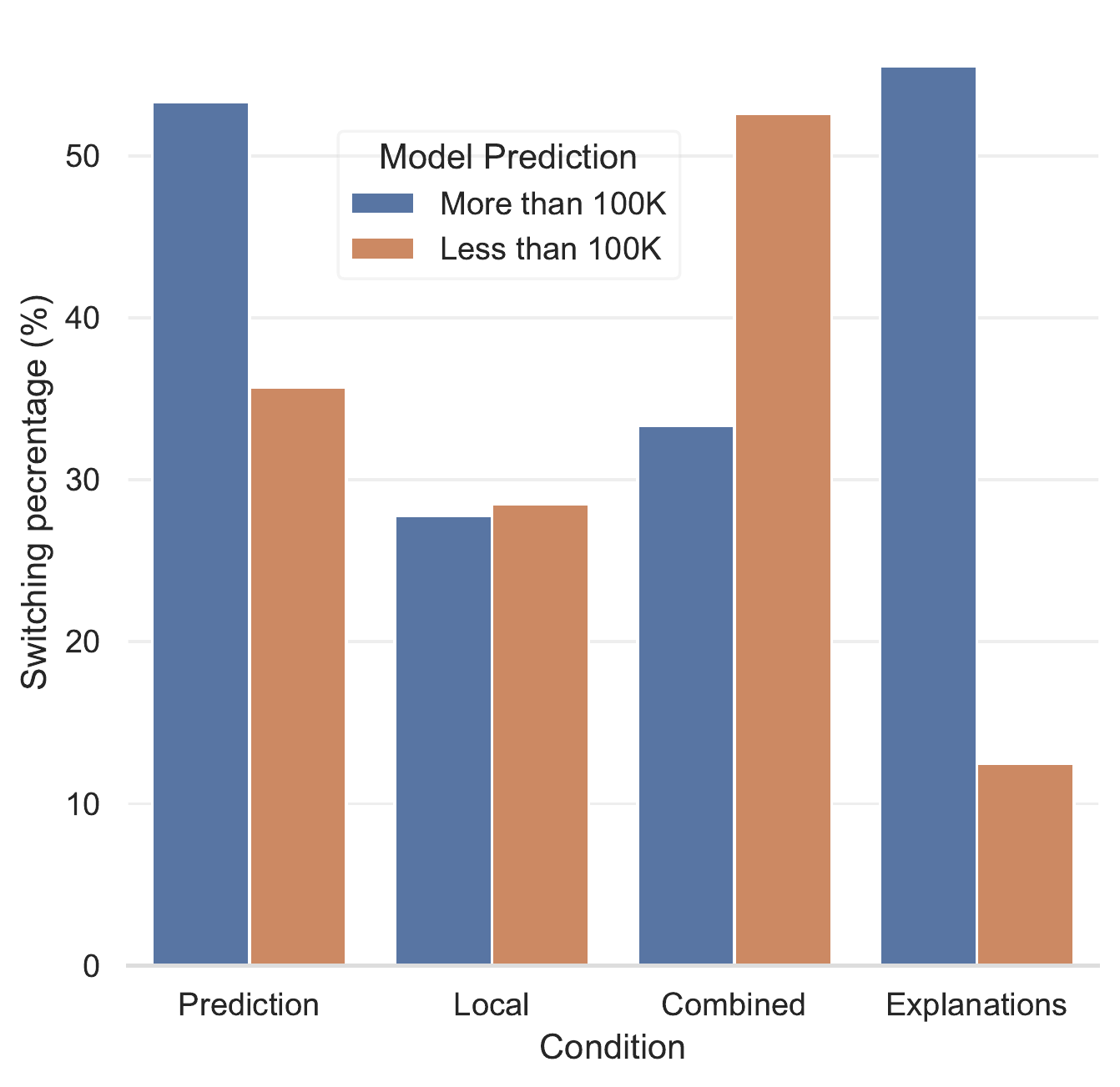}  
}
\caption{The switching percentages for the different model predictions. Each subplot corresponds to a combination of human and model confidence.}
\label{switches}
\end{figure}

\subsection{Switching and Agreement}\label{switchaggr}

We concluded our analysis by addressing two issues, starting with the effect of pairing uncertainty estimates of different scopes, and then moving to the potential pitfalls of utilizing switching and agreement percentages to measure trust. To this end, we began with a brief qualitative analysis of users' switching behaviour. Unfortunately, when isolating trials where participants' initial answer differed from the model's prediction, the statistical power of our analysis is greatly reduced, so our tests fail to detect significant differences. Despite that, there are some clear patterns present in the data, from which we can gain valuable insights, so we opted for providing a qualitative analysis, instead of dismissing them. Overall, participants' switching percentage in the \textbf{Prediction}, \textbf{Local}, \textbf{Combined}, \textbf{Explanations} conditions was $50\%$, $37\%$, $45\%$, and $34\%$, respectively. Furthermore, in all conditions switching helped participants improve their performance, since by altering their initial prediction to follow the model’s suggestion their accuracy increased by $41\%$, $25\%$, $15\%$, and $17\%$, following the same order as before.

Focusing on the \textbf{Local} and \textbf{Combined} conditions, we looked for differences in switching behaviour that can be explained by the fact that global error rates were available in the latter, but not in the former. Figure \ref{switches} depicts the percentage of trials participants switched their prediction, depending on Condition, Human Confidence, and Model Confidence, where we differentiate between cases where the model predicts Less than $100$K and those where it predicts More than $100$K. In the (Human - High \& Model - Low) combination participants exhibited a similar behaviour in both conditions, presumably because their behaviour was driven by their own intuitions. However, in every other confidence combination participants’ behaviour in the \textbf{Local} and \textbf{Combined} conditions were strikingly different. One the one hand, in the \textbf{Local} condition, switching percentages between the two classes were almost identical, but on the other hand, in the \textbf{Combined} condition, the switching percentage when the model's prediction was Less than $100$K was much higher than when the prediction was More than $100$K, consistent with the view that the poor global error rates of the More than $100$K class lessened the chances of participants switching to match the model’s prediction. Inversely, the great global error rates in the Less than $100$K class prompted participants to follow these suggestions.


This is more clearly demonstrated when (Human - Low \& Model - High), where knowing that the model had $91\%$ success rate when predicting Less than $100$K, encouraged participants in the \textbf{Combined} condition to switch in $89\%$ of the trials, compared to $60\%$ in the \textbf{Local} one. In line with this reasoning, when the prediction was More than $100$K, participants in the former condition were aware that model performance was relatively poor, so their switching percentage plummeted to $41\%$, which is substantially lower than the $57\%$  in the \textbf{Local} condition. This observation perfectly captures the added benefits of pairing these estimates together, as global error rates convey information about the robustness of local confidence scores themselves, which is in line with \textbf{H4}, however, additional studies are necessary in order to provide more robust evidence confirming this effect. 

\begin{figure}
    \centering
    \includegraphics[scale=0.55]{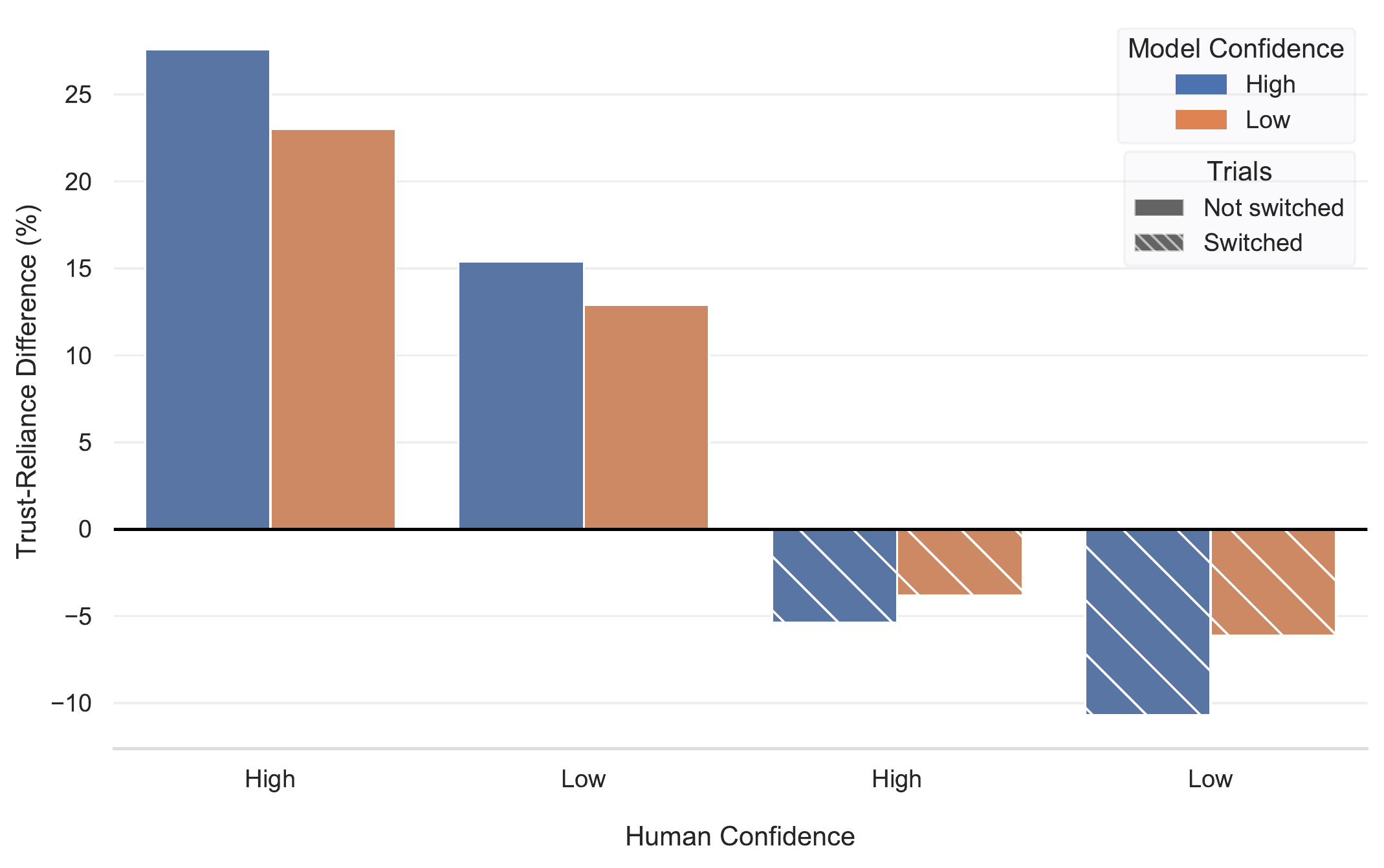}
    \caption{The difference between trust and reliance, in terms of the interaction of human and model confidence. Solid bars correspond to trials where participants did not switch their prediction, while dashed ones are computed based on switching trials.}
    \label{trustreliance}
\end{figure}

In the same vein, while the \textbf{Combined} and \textbf{Explanations} conditions followed a similar trend for instances with high human confidence, the pattern was drastically different for low confidence instances. Especially when (Human - Low \& Model - Low), the trends got reversed, which could be interpreted as additional evidence that explanations promoted case by case reasoning. According to this account, participants in the \textbf{Explanations} condition looked past the poor error rate of the More than $100$K predictions, using explanations to verify whether the model's reasoning was sound for the instance at hand. Notably they were very successful in doing so, since their accuracy in cases where they switched to follow a More than $100$K model prediction was $80\%$. Future research should investigate this topic in more detail, however this pattern along with the one in Section \ref{sec:trust}, provided some very promising indications in favour of this interpretation of the results.

Finally, we discuss a pattern that illustrates the non-equivalence of reliance and trust. Figure \ref{trustreliance} shows  the average difference between participants' trust and reliance scores, once considering trials where participants did not switch their predictions (regardless of whether they initially agreed with the model), and once considering only trials where they switched. In the former, there was a positive trend for all human/model combinations of confidence, meaning that participants' trust scores were higher than their reliance ones. However, when considering only switching trials, a stark contrast was observed, with the trend getting completely reversed, and  reliance scores dominating the corresponding trust ones. We should note that this discrepancy was induced by differences in reliance, since although participants' trust increased by $5.51\%$ in switching trials, the corresponding increase in reliance was equal to an impressive $33.11\%$. Even though we only offer a qualitative account of this phenomenon,  the observed pattern is consistent with previous works that argue that both agreement and switching percentages are indicators of reliance, not trust. Adding to this, we found that in $29\%$ of all trials where participants and model agreed, their reported reliance scores were lower that $3$ out of $7$, meaning that their predictions were predominantly driven by their own intuitions. This indicates that switching percentage is a stronger indicator of reliance, since human-model agreement on its own does not necessarily imply high reliance. Regardless, interpreting either as a manifestation of trust may result to misleading conclusions.

\section{Discussion}

In this section we discuss and contextualize our results, as well as we propose several future research directions.

\subsection{The role of human confidence}

Our findings provided strong evidence that human confidence has a major effect on multiple aspects of the joint human-AI synergy. Extending the results in \citep{bansal2021does}, we showed that humans were predominantly  benefited by the model's assistance in cases where they are uncertain, but the model made high confidence predictions. This finding is in line with highly influential existing theories on human-computer interaction \citep{lee2004trust,hoff2015trust}, where it is argued that users' self-confidence impacts their attitude  towards automation. Furthermore, the results presented in Section \ref{sec:perf}, demonstrated  that not accounting for human confidence may severely distort an analysis. In light of these findings, future experimental studies should be designed in a way that records or controls for human confidence, instead of solely focusing on model confidence. Interestingly, an emerging line of research calls for training ML models using procedures that incorporate human confidence \citep{bansal2021most,mozannar2020consistent,wilder2020learning}, indicating that there is a general interest into utilizing and accounting for this factor. 

Beyond  predictive performance, our findings suggested that the influence of human confidence extends to users' reliance, understanding, and trust towards a model. Moreover the discussion in Section \ref{switchaggr}, emphasized that human confidence also influenced switching and agreement percentages, while also raising concerns about the suitability of these two indicators to assess trust. Previous research has been consistent that both of these measure reliance \citep{dixon2006automation,madhavan2010effects,miller2016behavioral}, and has discussed the caveats of studying trust through reliance \citep{chancey2015role,hussein2020trust}. In our opinion, this calls for rethinking experimental designs or for adjusting the way final results are interpreted. A potential resolution would be to compliment reliance indicators with items from specialized trust measuring scales, and assess trust based on both, which has been the standard practice within the human factors and human-computer interaction communities \citep{wang2009trust,chancey2013role,moray2000adaptive,merritt2008not}, or to use more elaborated behavioural indicators that capture multiple aspects of trust, such as those in \citep{de2003effects,miller2016behavioral}. An alternative to modifying the experimental designs, would be to motivate surveys and form hypotheses in terms of reliance  \citep{lee2004trust}.

\subsection{The complementary effect of uncertainty and explanations}

Another central question we explored in this work concerns the role of combining uncertainty estimates and explanations. Prior work suggested that in terms of accuracy, pairing model predictions with the corresponding confidence is as effective as pairing them with explanations \citep{bansal2021does,lai2019human,lai2020chicago}, implying that, performance-wise, uncertainty estimates are as powerful as explanations, while arguably being simpler to understand and implement. Consistent with this idea, our results provided evidence that when both predictions and confidence information were available, providing participants  with additional information did not lead to better performance. Despite that, we identified a strong complementary effect, since participants in the \textbf{Explanations} condition had significantly higher self-reported understanding, while also exhibiting a far superior objective model understanding. Interestingly, although only feature importance explanation were provided, their effect permeated multiple aspects of model understanding. Increased understanding has been linked to higher rates of model acceptance \citep{shin2021effects}, while the findings in \citep{ashoori2019ai} indicate that when the stakes are high, ethical considerations may lead to people entirely dismissing a model, regardless of its accuracy, unless they are able to understand its decision-making process. A promising future direction is to adopt a longitudinal experimental design and quantify the effect of explanations on model acceptance or retention. In general, user behaviour is shaped over multiple interactions with the model through an extended period of time, where unexpected or otherwise surprising behaviour may manifest, so longitudinal designs have the potential to provide important insights that are missed by cross sectional designs, which do not record how user behaviour changes over extended periods of time.

Moreover, our results indicated that complementary effects can be found within uncertainty measures too, as discussed in Section \ref{switchaggr}. This is consistent with the recent discussions in \citep{bhatt2021uncertainty}, demonstrating how communicating different kinds of uncertainty information can induce different user behaviour. In this work we considered predicted probabilities and recall, however there is a lot of room for exploring different measures or combinations thereof, such as precision, false discovery rate, etc. In particular, we find the approach of combining information with diverse scopes (e.g. local and global) to be very promising and worthy of further exploration. An immediate follow up study stemming from our work could explore the effect of more refined global uncertainty information. For example, instead of providing the overall recall of each class, we could first cluster the datapoints based on similarity, and then compute cluster-wise recalls. This localized version of a global summary allows for capturing potential variability in model performance within the same class, depending on sub-population characteristics. However, it should be noted that such approaches require users to have a certain level of numerical competency, which differs substantially from person to person \citep{zikmund2007validation}, so alternatives exploring visualizations and/or natural language expressions of uncertainty should be considered as well.

\subsection{Explanations in AI}

Our findings suggested that explanations provided unique insights that impact model understanding, however explanatory needs are highly dependent on the application \citep{zhou2021evaluating,ribera2019can}. Our work only considered feature importance explanations, however alternative scenarios may call for different types of explanations, such as generating counterfactual instances \citep{counterfactual} or propositional rules \citep{AAAI1816982}. Although there is a number of recent surveys that compare the effect of various explanation types \citep{wang2021explanations,bansal2021does,lai2019human}, to our knowledge there has not been a systematic effort to study the relationship between application characteristics and explanation style preference or efficacy. Furthermore, even within the same application, we expect stakeholders of different expertise to have different explanatory preferences.

Finally, in Section \ref{sec:perf}, we provided evidence that when participants had low confidence, model assistance significantly improved their performance, especially when the model generated high confidence predictions. Having said that, when both parties had high confidence, we mostly observed a downwards trend, which resulted in a significant decline in performance in the \textbf{Explanations} condition. It is possible that this finding was due to participants' having an information overload \citep{poursabzi2021manipulating}, where they had a hard time keeping track of all the information that was presented to them. However, other surveys have raised concerns about human over-reliance on a model when explanations are provided \citep{bansal2021does,kaur2020interpreting}, so the observed decline in accuracy might be related to this phenomenon. In our view, a promising step towards resolving this situation could be to explore the effect of communicating information about the robustness of an explanation. Most XAI techniques heavily rely on approximations, which means that the final explanation might not be faithful to the model, thus distorting its decision-making process. Moreover, even if no approximations are performed, explanations might face stability issues, where small feature perturbations may lead to drastically different explanations \citep{yeh2019fidelity}. If presented with such information, it is reasonable to assume that users would be more skeptical of explanations, thus reducing their over-reliance. All things considered, we believe that the interplay between uncertainty and explanations calls for further exploration, as it can be integral in guiding the safe and responsible adaptation of automated systems.

\section{Limitations}

We acknowledge that one limitation of our study is that we only recruited participants residing in USA, thus we make no claims about the cross-cultural validity of our results. Moreover, we did not record information about participants' familiarity and attitude towards AI, so our results may be influenced by participants predispositions towards automation. Furthermore, participants were not experts on salary prediction tasks. We alleviated this limitation by including a familiarization phase in our experiment. The fact that participants' performance was comparable to the model's indicates that our approach was effective. 

Another limitation is that participants were not held liable for their performance, which bared no consequence to them. We addressed this limitation by providing additional performance-based rewards to motivate participants strive for optimal performance.

\section{Conclusions}

Previous empirical studies have demonstrated that pairing model predictions and confidence is more effective than explanations in assisting humans improve their accuracy in decision-making tasks. In this work we ask whether bringing them together can provide complementary, non-accuracy related benefits, while also exploring how the interaction of human and model confidence influences human-AI joint accuracy, reliance, understanding, and trust towards the model. To this end, we conducted a study with $112$ human participants. We found strong evidence suggesting that human performance is improved in cases where they have low confidence themselves, but the model makes high confidence predictions. Moreover, we found that pairing uncertainty estimates with explanations induces a complementary effect, resulting in high performance and significantly better model understanding. We concluded our findings by providing a qualitative analysis outlining the benefits of combining uncertainty estimates with different scopes, as well as the potential pitfalls of utilizing reliance indicators to measure trust.

We hope that this work will motivate future research that further investigates the role of self-confidence and how different combinations of information influence the human-AI collaboration, in situations where time constraints or other inherent risks are present. Furthermore, another promising direction would be to explore  whether interactive methods where humans can actively enquiry a model to satisfy their explanatory needs yield additional benefits, compared to static strategies (like the ones considered in this experiment). Achieving a synergistic relationship between humans and AI is set to be one of the main end goals of the responsible incorporation of AI in our society, and advances along these lines should hopefully bring us a step closer to achieving these endeavours. 

\begin{ack}
We would like to thank Peter Gostev for all the stimulating discussions, which heavily contributed into pursuing the research questions considered in this work. Moreover, we are grateful to Maria Mavridaki for her feedback and suggestions, which greatly improved the quality of the final manuscript.
\end{ack}

\bibliographystyle{plainnat}
\bibliography{ref}








\clearpage

\appendix

\section{CIs and Comparisons}\label{effects}

Here are all the details of the CIs and comparisons that were presented in the main paper. All CIs are open, while CIs and p-values have been adjusted using the Bonferroni correction method in order to control the family-wise error rate.

\subsection{CIs for Section \ref{sec:perf}}

\begin{table}[h]
\centering
\scalebox{1}{\begin{tabular}{|l|c|c|}
\hline
 Condition & Average Accuracy & $95\%$ CI\\
\hline  
\textbf{Prediction} & $70.5\%$ & $(64.7, 76.7)$ \\
\textbf{Local}  & $74.1\%$ &  $(69.8, 79.4)$\\
\textbf{Combined} &  $77.6\%$ & $(73.4, 81.6)$\\
\textbf{Explanations} & $76.7\%$ & $(73.6, 79.9)$\\
\hline
\end{tabular}}
\caption{Participants' unassisted  accuracy}
\label{unassisted}
\end{table}

\begin{table*}[h]
\centering
\scalebox{1}{\begin{tabular}{|l|c|c|}
\hline
 Condition & Average Accuracy & $95\%$ CI\\
\hline  
\textbf{Prediction} & $77.9\%$ & $(74.7, 80.8)$ \\
\textbf{Local}  & $78.5\%$ &  $(75, 82.3)$\\
\textbf{Combined} &  $79.9\%$ & $(76.3, 83.4)$\\
\textbf{Explanations} & $78.3\%$ & $(75.4, 81)$\\
\hline
\end{tabular}}
\caption{Participants' assisted  accuracy}
\label{assisted}
\end{table*}

\begin{table*}[h]
\centering
\scalebox{1}{\begin{tabular}{|l|c|c|}
\hline
 Condition & Average Difference & $95\%$ CI\\
\hline  
\textbf{Prediction} & $7.36\%$ & $(2.45, 11.8) $ \\
\textbf{Local}  & $4.46\%$ &  $(0.44, 7.81)$\\
\textbf{Combined} &  $2.23\%$ & $(-1.11, 5.35)$\\
\textbf{Explanations} & $1.56\%$ & $(-1.33, 4.68)$\\
\hline
\end{tabular}}
\caption{Difference in participants' assisted and unassisted accuracy}
\label{diffacc}
\end{table*}

\begin{table*}[h]
\centering
\scalebox{1}{\begin{tabular}{|l|c|c|c|}
\hline
 Condition & Model Confidence & Average Difference & $95\%$ CI\\
\hline  
\textbf{Prediction} & Low & $1.33\%$ & $(-4.46, 6.25) $ \\
\textbf{Prediction} & High & $13.39\%$ & $(2.67, 22.32) $ \\
\textbf{Local} & Low & $4.91\%$ &  $(-2.67, 10.71)$\\
\textbf{Local} & High & $4.01\%$ &  $(-1.78, 9.82)$\\
\textbf{Combined} & Low &  $3.57\%$ & $(-3.12, 8.92)$\\
\textbf{Combined}& High &  $0.89\%$ & $(-4.46, 6.25)$\\
\textbf{Explanations} & Low & $1.78\%$ & $(-2.23, 5.8)$\\
\textbf{Explanations} & Low & $1.33\%$ & $(-2.67, 5.8)$\\
\hline
\end{tabular}}
\caption{Difference in participants' assisted and unassisted accuracy, with respect to the levels of model confidence}
\label{diffaccmodel}
\end{table*}

\begin{table}[H]
\centering
\scalebox{1}{\begin{tabular}{|l|c|c|c|c|}
\hline
 Condition & Human Confidence & Model Confidence & Average Difference & $95\%$ CI\\
\hline  
\textbf{Prediction} & Low & Low & $5.35\%$ & $(-2.67, 1.25) $ \\
\textbf{Prediction} & High & Low & $-2.67\%$ & $(-8.92, 3.57) $ \\
\textbf{Prediction} & Low  & High & $22.32\%$ & $(10.71, 33.03) $ \\
\textbf{Prediction} & High  & High & $4.46\%$ & $(-7.14, 14.28) $ \\
\textbf{Local} & Low & Low & $8.92\%$ &  $(-1.78, 17.85)$\\
\textbf{Local} & High & Low & $0.89\%$ &  $(-5.35, 6.25)$\\
\textbf{Local} & Low & High & $10.71\%$ &  $(2.67, 17.85)$\\
\textbf{Local} & High & High & $-2.67\%$ &  $(-8.03, 2.67)$\\
\textbf{Combined} & Low & Low &  $8.92\%$ & $(0.89, 16.07)$\\
\textbf{Combined} & High & Low &  $-1.78\%$ & $(-8.92, 4.46)$\\
\textbf{Combined} & Low & High &  $7.14\%$ & $(-2.67, 16.96)$\\
\textbf{Combined} & High & High &  $-5.35\%$ & $(-10.71, 0.89)$\\
\textbf{Explanations} & Low & Low & $6.25\%$ & $(-2.67, 13.39)$\\
\textbf{Explanations} & High & Low & $-2.67\%$ & $(-5.35, 1.78)$\\
\textbf{Explanations} & Low & High & $8.03\%$ & $(0.0, 16.07)$\\
\textbf{Explanations} & High & High & $-5.35\%$ & $(-9.82, 0.0)$\\
\hline
\end{tabular}}
\caption{Difference in participants' assisted and unassisted accuracy, with respect to the levels of human and model confidence}
\label{diffaccmodelhuman}
\end{table}

\subsection{Effects and comparisons for Section \ref{sec:rel}}

\begin{table}[H]
\centering
\scalebox{1}{\begin{tabular}{|l|c|c|c|}
\hline
 Factor & df & Statistic & p-value\\
\hline  
Condition & $3$ & $5.529 $ & $0.137$\\
Model Confidence    & $1$ &  $5.138$ & $0.023$\\
Condition$\times$Model Confidence  &  $3$ & $17.574$ & $0.001$\\
Human Confidence  & $1$ & $40.17$ & $<0.001$\\
Condition$\times$Human Confidence & $3$ & $5.255$ & $0.154$\\
Model Confidence$\times$Human Confidence  & $1$ & $1.344$ & $0.246$\\
Condition$\times$Model Confidence$\times$Human Confidence  & $3$ & $2.703$ & $0.44$\\
\hline
\end{tabular}}
\caption{ANOVA table for Section \ref{sec:rel}}
\label{diffrel}
\end{table}

\begin{table}[H]
\centering
\scalebox{1}{\begin{tabular}{|l|c|c|c|}
\hline
 Condition & Average Difference & Statistic & p-value\\
\hline  
\textbf{Prediction} & $0.15\%$ & $198$ & $1$ \\
\textbf{Local}  & $9.87\%$ &  $32$ & $<0.001$\\
\textbf{Combined} &  $1.41\%$ & $166$ & $1$\\
\textbf{Explanations} & $-1.65\%$ & $175$ & $1$\\
\hline
\end{tabular}}
\caption{Difference in participants' reliance between high and low confidence model predictions}
\label{diffcondrel}
\end{table}

\subsection{Effects and comparisons for Section \ref{sec:understannding1}}

\begin{table}[H]
\centering
\scalebox{1}{\begin{tabular}{|l|c|c|c|}
\hline
 Factor & df & Statistic & p-value\\
\hline  
Condition & $3$ & $10.944 $ & $0.012$\\
Model Confidence    & $1$ &  $23.015$ & $<0.001$\\
Condition$\times$Model Confidence  &  $3$ & $6.765$ & $0.08$\\
Human Confidence  & $1$ & $18.114$ & $<0.001$\\
Condition$\times$Human Confidence & $3$ & $2.206$ & $0.531$\\
Model Confidence$\times$Human Confidence  & $1$ & $3.963$ & $0.047$\\
Condition$\times$Model Confidence$\times$Human Confidence  & $3$ & $4.788$ & $0.188$\\
\hline
\end{tabular}}
\caption{ANOVA table for Section \ref{sec:understannding1}}
\label{diffund}
\end{table}

\begin{table}[H]
\centering
\scalebox{1}{\begin{tabular}{|l|c|c|c|}
\hline
 Human/Model Confidence Contrast & Average Difference & Statistic & p-value\\
\hline  
High/High vs High/Low & $6.41\%$ & $1555$ & $<0.001$ \\
High/High vs Low/Low  & $7.84\%$ &  $1148.5$ & $<0.001$\\
High/High vs Low/High &  $4.72\%$ & $1755$ & $<0.001$\\
High/Low vs Low/Low & $1.69\%$ & $2683$ & $0.975$\\
High/Low vs Low/High & $-1.42\%$ & $2848$ & $1$\\
Low/High vs Low/Low & $3.12\%$ & $2443$ & $0.218$\\
\hline
\end{tabular}}
\caption{Difference in participants' understanding between the various configurations of human/model confidence}
\label{diffhumanmodelund}
\end{table}

\begin{table}[H]
\centering
\scalebox{1}{\begin{tabular}{|l|c|c|c|}
\hline
 Condition Contrast & Average Difference & Statistic & p-value\\
\hline  
\textbf{Explanations} vs \textbf{Prediction} & $5.58\%$ & $-1.13$ & $0.5$ \\
\textbf{Explanations} vs \textbf{Local}  & $7.83\%$ &  $-2.5$ & $0.036$\\
\textbf{Explanations} vs \textbf{Combined} &  $7.97\%$ & $-3.01$ & $0.007$\\
\hline
\end{tabular}}
\caption{Difference in participants' subjective understanding between \textbf{Explanations} and the remaining conditions}
\label{diffcondund}
\end{table}

\subsection{Effects and comparisons for Section \ref{sec:trust}}

\begin{table}[H]
\centering
\scalebox{1}{\begin{tabular}{|l|c|c|c|}
\hline
 Factor & df & Statistic & p-value\\
\hline  
Condition & $3$ & $1.862 $ & $0.601$\\
Model Confidence    & $1$ &  $12.942$ & $<0.001$\\
Condition$\times$Model Confidence  &  $3$ & $14.817$ & $0.002$\\
Human Confidence  & $1$ & $46.269$ & $<0.001$\\
Condition$\times$Human Confidence & $3$ & $0.661$ & $0.882$\\
Model Confidence$\times$Human Confidence  & $1$ & $1.358$ & $0.244$\\
Condition$\times$Model Confidence$\times$Human Confidence  & $3$ & $2.078$ & $0.556$\\
\hline
\end{tabular}}
\caption{ANOVA table for Section \ref{sec:trust}}
\label{difftrust}
\end{table}
\clearpage

\begin{table}[H]
\centering
\scalebox{1}{\begin{tabular}{|l|c|c|c|}
\hline
 Condition & Average Difference & Statistic & p-value\\
\hline  
\textbf{Prediction} & $-1.64\%$ & $153$ & $1$ \\
\textbf{Local}  & $6.34\%$ &  $88$ & $0.035$\\
\textbf{Combined} &  $2.1\%$ & $160$ & $1$\\
\textbf{Explanations} & $8.37\%$ & $77$ & $0.016$\\
\hline
\end{tabular}}
\caption{Difference in participants' trust between high and low confidence model predictions}
\label{diffcondtrust}
\end{table}

\subsection{CIs for Section \ref{objectiveunder}}

\begin{table}[H]
\centering
\scalebox{1}{\begin{tabular}{|l|c|c|c|}
\hline
Contrast & Aspect &  Average Difference & $95\%$ CI\\
\hline  
\textbf{Local} vs \textbf{Prediction} & Global feature importance  & $-0.143$ & $(-0.35, 0.1) $ \\
\textbf{Combined} vs \textbf{Prediction} & Global feature importance & $0$ & $(-0.28, 0.28) $ \\
\textbf{Explanations} vs \textbf{Prediction} & Global feature importance  &  $0.964$ & $(0.64, 1.28) $ \\
\textbf{Local} vs \textbf{Prediction} & Local feature importance  &  $0$ & $(0,0) $ \\
\textbf{Combined} vs \textbf{Prediction} & Local feature importance &  $0$ &  $(0,0)$\\
\textbf{Explanations} vs \textbf{Prediction} & Local feature importance  & $0.643\%$ &  $(0.46, 0.82)$\\
\textbf{Local} vs \textbf{Prediction} & Counterfactuals  & $-0.036$ &  $(-0.42, 0.35)$\\
\textbf{Combined} vs \textbf{Prediction} & Counterfactuals  & $-0.143$ &  $(-0.46, 0.17)$\\
\textbf{Explanations} vs \textbf{Prediction} & Counterfactuals &  $0.893$ & $(0.53, 1.25)$\\
\textbf{Local} vs \textbf{Prediction} & Model simulation &  $0.25$ & $(-0.35, 0.85)$\\
\textbf{Combined} vs \textbf{Prediction} & Model simulation &   $0.143$ & $(-0.32, 0.60)$\\
\textbf{Explanations} vs \textbf{Prediction} & Model simulation & $0.5$ & $(0.07, 0.92)$\\
\textbf{Local} vs \textbf{Prediction} & Error detection &  $0.25$ & $(-0.25, 0.75)$\\
\textbf{Combined} vs \textbf{Prediction} & Error detection & $0.036$ & $(-0.39, 0.46)$\\
\textbf{Explanations} vs \textbf{Prediction} & Error detection & $-0.036$ & $(-0.5, 0.42)$\\
\hline
\end{tabular}}
\caption{Difference in participants' objective model understanding}
\label{contraststest}
\end{table}
\clearpage
\section{Objective Model Understanding Questions}\label{questionstrust}

The following are the $9$ multiple choice questions that were used to assess participants' objective model understanding. The qualitative aspects of all questions and answers agree with the ones found in the test developed in \citep{wang2021explanations}, except from the fact that we do not allow for multiple correct answers. The correct answers are in  \textcolor{red}{red}.

\subsection{Global Feature Importance}
\subsubsection{Question 1}

In general, the value of which feature has the greatest influence on the model's predictions?
\begin{center}
\begin{minipage}{.5\textwidth}

    \begin{tabenum}
  \tabenumitem[a)] Age 
  \tabenumitem[b)] Employer
  
  \tabenumitem[c)] Education
  \tabenumitem[\textcolor{red}{d)}] \textcolor{red}{Marital Status}
  
  \tabenumitem[e)] Occupation
  \tabenumitem[f)] Ethnic Background

  \tabenumitem[g)] Gender
  \tabenumitem[h)] Hours-per-week
  \end{tabenum}
  \end{minipage}
\end{center}

\subsubsection{Question 2}

In general, the value of which feature has the least influence on the model's predictions?
\begin{center}
\begin{minipage}{.5\textwidth}

    \begin{tabenum}
  \tabenumitem[a)] Age 
  \tabenumitem[b)] Employer
  
  \tabenumitem[c)] Education
  \tabenumitem[d)] Marital Status
  
  \tabenumitem[e)] Occupation
  \tabenumitem[\textcolor{red}{f)}] \textcolor{red}{Ethnic Background}

  \tabenumitem[g)] Gender
  \tabenumitem[h)] Hours-per-week
  \end{tabenum}
  \end{minipage}
\end{center}

\subsection{Local Feature Importance}

\subsubsection{Question 3}

\begin{figure}[H]
    \centering
    \includegraphics[scale=0.65]{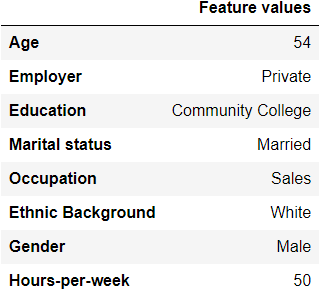}
    \caption{Question 3}
    \label{fig:my_label87}
\end{figure}

For this particular person, the value of which feature had the greatest influence on the model's prediction?

\begin{center}
\begin{minipage}{.5\textwidth}

    \begin{tabenum}
  \tabenumitem[a)] Age 
  \tabenumitem[b)] Employer
  
  \tabenumitem[c)] Education
  \tabenumitem[\textcolor{red}{d)}] \textcolor{red}{Marital Status}
  
  \tabenumitem[e)] Occupation
  \tabenumitem[f)] Ethnic Background

  \tabenumitem[g)] Gender
  \tabenumitem[h)] Hours-per-week
  \end{tabenum}
  \end{minipage}
\end{center}

\clearpage

\subsection{Counterfactuals}

\subsubsection{Question 4}
\begin{figure}[H]
    \centering
    \includegraphics[scale=0.65]{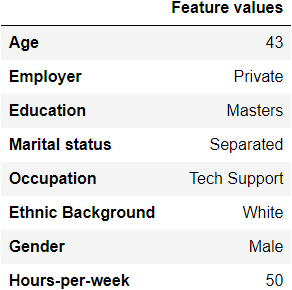}
    \caption{Question 4}
    \label{fig:my_label65}
\end{figure}

Our model currently predicts this person earns more than 100K dollars. If we change only one feature of this profile but leave all other features unchanged, which of the following changes is going to change our model’s prediction (i.e., make the model predict that the person earns less than 100K dollars)? 

\begin{center}
\begin{minipage}{.5\textwidth}

    \begin{tabenum}
  \tabenumitem[\textcolor{red}{a)}] \textcolor{red}{Change Age from 43 to 25}
  
  \tabenumitem[b)] Change Marital Status from Separated to Married
  
  \tabenumitem[c)] Change Ethnic Background from White to Black
  
  \tabenumitem[d)] Change Gender from Male to Female
  \end{tabenum}
  \end{minipage}
\end{center}
\clearpage

\subsubsection{Question 5}
\begin{figure}[H]
    \centering
    \includegraphics[scale=0.65]{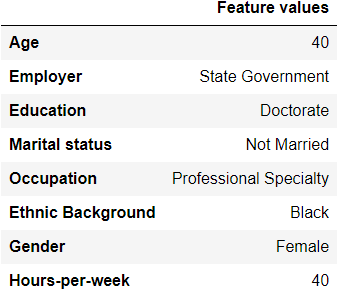}
    \caption{Question 5}
    \label{fig:my_label655}
\end{figure}

Our model currently predicts this person earns less than 100K dollars. If we change only one feature of this profile but leave all other features unchanged, which of the following changes is going to change our model’s prediction (i.e., make the model predict that the person earns more than 100K dollars)? 

\begin{center}
\begin{minipage}{.5\textwidth}

    \begin{tabenum}
  \tabenumitem[a)] Change Age from 40 to 50
  
  \tabenumitem[b)] Change Employer from State Government to Federal Government
  
  \tabenumitem[\textcolor{red}{c)}] \textcolor{red}{Change Marital Status from Not Married to Married}
  
  \tabenumitem[d)] Change Hours-per-week from 40 to 45
  \end{tabenum}
  \end{minipage}
\end{center}
\clearpage

\subsection{Model Simulation}
\subsubsection{Question 6}
\begin{figure}[H]
    \centering
    \includegraphics[scale=0.65]{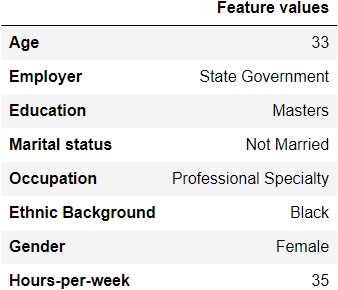}
    \caption{Question 6}
    \label{fig:my_label655}
\end{figure}

What do you think our model will predict for this person?

\begin{center}
\begin{minipage}{.5\textwidth}

    \begin{tabenum}
  \tabenumitem[\textcolor{red}{a)}] \textcolor{red}{The model will predict this person earns Less than 100K dollars}
  
  \tabenumitem[b)] The model will predict this person earns More than 100K dollars
  \end{tabenum}
  \end{minipage}
\end{center}

\subsubsection{Question 7}
\begin{figure}[H]
    \centering
    \includegraphics[scale=0.65]{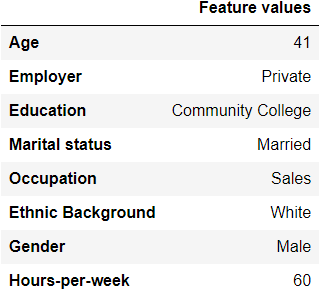}
    \caption{Question 7}
    \label{fig:my_label655}
\end{figure}

What do you think our model will predict for this person?

\begin{center}
\begin{minipage}{.5\textwidth}

    \begin{tabenum}
  \tabenumitem[a)] The model will predict this person earns Less than 100K dollars
  
  \tabenumitem[\textcolor{red}{b)}] \textcolor{red}{The model will predict this person earns More than 100K dollars}
  \end{tabenum}
  \end{minipage}
\end{center}

\subsection{Error Detection}

\subsubsection{Question 8}
\begin{figure}[H]
    \centering
    \includegraphics[scale=0.65]{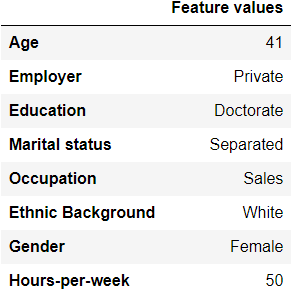}
    \caption{Question 8}
    \label{fig:my_label655}
\end{figure}

Our model predicts that this person earns Less than 100K dollars. Do you believe this prediction is correct?

\begin{center}
\begin{minipage}{.5\textwidth}

    \begin{tabenum}
  \tabenumitem[a)] Yes, I think this prediction is correct
  
  \tabenumitem[\textcolor{red}{b)}] \textcolor{red}{No, I think this prediction is wrong}
  
  \end{tabenum}
  \end{minipage}
\end{center}

\subsubsection{Question 9}
\begin{figure}[H]
    \centering
    \includegraphics[scale=0.65]{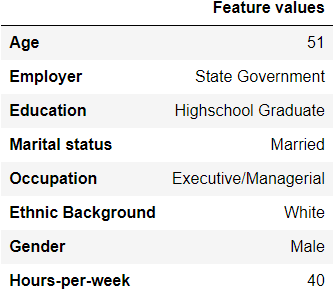}
    \caption{Question 9}
    \label{fig:my_label655}
\end{figure}

Our model predicts that this person earns Less than 100K dollars. Do you believe this prediction is correct?

\begin{center}
\begin{minipage}{.5\textwidth}

    \begin{tabenum}
  \tabenumitem[a)] Yes, I think this prediction is correct
  
  \tabenumitem[\textcolor{red}{b)}] \textcolor{red}{No, I think this prediction is wrong}
  
  \end{tabenum}
  \end{minipage}
\end{center}

\end{document}